\documentclass[5p]{elsarticle}
\usepackage{bm}
\usepackage{hyperref}

\usepackage{siunitx,tabu}
\usepackage{graphicx}
\usepackage{caption}
\usepackage{subcaption}
\usepackage{natbib}
\usepackage{color}
\usepackage{makecell}
\usepackage[english]{babel}

\usepackage{multirow, rotating, array}
\usepackage{amsfonts, amssymb, amsmath}
\usepackage{enumitem}
\usepackage{pdflscape}

\usepackage{floatrow}
\usepackage{flafter}

\usepackage{color, graphicx, float}
\usepackage{subcaption}
\usepackage{afterpage}
\usepackage{flafter}
\usepackage{algorithmic}
\usepackage[algoruled]{algorithm2e}

\usepackage{wrapfig}
\usepackage{makeidx}
\usepackage{makecell}
\usepackage{booktabs}
\usepackage[normalem]{ulem}
\newcommand{\thickhline}{%
	\noalign {\ifnum 0=`}\fi \hrule height 1pt
	\futurelet \reserved@a \@xhline
}

\journal{Medical Image Analysis}

\hypersetup{
	colorlinks   = true, %
	urlcolor     = blue, %
	linkcolor    = red, %
	citecolor    = red   %
}

\biboptions{authoryear}

\newcommand{\reviewminortwo}[1]{\textcolor{magenta}{#1}}
\newcommand{\reviewminor}[1]{\textcolor{magenta}{#1}}
\newcommand{\review}[1]{\textcolor{magenta}{#1}}
\renewcommand{\review}[1]{#1}
\renewcommand{\reviewminor}[1]{#1}
\renewcommand{\reviewminortwo}[1]{#1}
\renewcommand{\sout}[1]{}

\begin{document}

\begin{frontmatter}

%\title{Identification of patients with functionally significant coronary artery stenosis: A deep learning analysis of left ventricular myocardium in coronary CT angiography}

%\title{A deep learning analysis of the left ventricular myocardium for identification of patients with functionally significant coronary artery stenosis using coronary CT angiography}

\title{Deep learning analysis of the myocardium in coronary CT angiography for identification of patients with functionally significant coronary artery stenosis}

\author[add1]{Majd Zreik}

\ead{m.zreik@umcutrecht.nl}
\author[add1]{Nikolas Lessmann}

\author[add2]{Robbert W. van Hamersvelt}

\author[add1]{Jelmer M. Wolterink}

\author[add3]{Michiel Voskuil}

\author[add1]{Max A. Viergever}

\author[add2]{Tim Leiner}

\author[add1]{Ivana I\v sgum}

\address{This paper was submitted in April 2017 and accepted in November 2017 for publication in Medical Image Analysis.
	
	Please cite as: Zreik et al., Medical Image Analysis, 2018, vol. 44, pp. 72-85.

}

\address[add1] {Image Sciences Institute, University Medical Center Utrecht and Utrecht University, Utrecht, The Netherlands}
\address[add2]{Department of Radiology, University Medical Center Utrecht and Utrecht University, Utrecht, The Netherlands}
\address[add3] {Department of Cardiology, University Medical Center Utrecht and Utrecht University, Utrecht, The Netherlands}

\begin{abstract}
In patients with coronary artery stenoses of intermediate severity, the functional significance needs to be determined. Fractional flow reserve (FFR) measurement, performed during invasive coronary angiography (ICA), is most often used in clinical practice. To reduce the number of ICA procedures, \sout{several methods use the non-invasive coronary CT angiography (CCTA) to determine the functional significance of a coronary stenosis. Such methods strongly rely on the geometry of the coronary artery tree. In contrast to these approaches} we present a method for automatic identification of patients with\sout{one or multiple} functionally significant coronary artery stenoses, employing deep learning analysis of the left ventricle (LV) myocardium in \review{rest coronary CT angiography} (CCTA).

The study includes consecutively acquired CCTA scans of 166 patients who underwent invasive FFR measurements. To \sout{automatically}identify patients with a functionally significant coronary artery stenosis, analysis is performed in \review{several} stages. First, the LV myocardium is segmented using a multiscale convolutional neural network (CNN).\sout{ Thereafter,the segmented myocardium is represented with a number of encodings generated by a} \review{To characterize the segmented LV myocardium, it is subsequently encoded using unsupervised} convolutional autoencoder (CAE).\sout{Since}\sout{It is expected that ischemic changes appear locally} \review{As ischemic changes are expected to appear locally,} the LV myocardium is divided into a number of spatially connected clusters, \review{and statistics of the encodings are computed as features}.\sout{To capture local variation in the appearance of the LV myocardium, the standard deviation of each encoding is computed over the voxels in each cluster. The standard deviations maxima are subsequently used to classify patients into those with functionally significant stenosis and those without it. Classification is performed using a support vector machine classifier. The invasive FFR measurements provided the reference for presence of a functionally significant stenosis. Forty CCTA scans were used to train and test the LV myocardium segmentation, and twenty CCTA scans were used to train and validate the myocardium encoding. Classification of patients was evaluated using the remaining 126 CCTA scans in 50 10-fold cross-validation experiments, where in every experiment patients were randomly assigned to training and testing sets.} \review{Thereafter, patients are classified according to the presence of functionally significant stenosis using an SVM classifier based on the extracted features.} 

\review{Quantitative evaluation of} LV myocardium segmentation \review{in 20 images} resulted in \review{an average}\sout{a} Dice coefficient of $0.91$ and \review{an average} mean absolute distance between the segmented and reference LV boundaries of $0.7$ mm.\sout{Classification of patients resulted in} \review{Twenty CCTA images were used to train the LV myocardium encoder. Classification of patients was evaluated in the remaining 126 CCTA scans in 50 10-fold cross-validation experiments and resulted in} an area under the receiver operating characteristic curve of $0.74 \pm 0.02$. At sensitivity levels 0.60, 0.70 and 0.80, the corresponding specificity was 0.77, 0.71 and 0.59, respectively.

The results demonstrate that automatic analysis of the LV myocardium in a single CCTA scan \review{acquired at rest}, without assessment of the anatomy of the coronary arteries, can be used to identify patients with functionally significant coronary artery stenosis. This might reduce the number of patients undergoing unnecessary invasive FFR measurements. 

\end{abstract}

\begin{keyword} 
 Functionally significant coronary artery stenosis \sep Convolutional autoencoder\sep Convolutional neural network\sep Fractional flow reserve \sep Coronary CT angiography\sep Deep learning 

\end{keyword}

\end{frontmatter}

\section{Introduction}

Obstructive coronary artery disease (CAD) is the most common type of heart disease \citep{AHA15}. Obstructive CAD occurs when one or more of the coronary arteries which supply blood to myocardium are narrowed owing to plaque buildup on the arteries' inner walls, causing stenosis. Only functionally significant stenoses, i.e stenoses that significantly limit the blood flow to the myocardium and lead to myocardial ischemia need to be treated to reduce CAD morbidity \citep{Pijl96,Toni09,Pijl10,Nune15}. Conversely, treating stenoses that are not functionally significant leads to more harm than benefit \citep{Pijl10,Pijl13}. Therefore, it is important to assess the severity of coronary stenoses with respect to their impact on the myocardium. 

To establish the functional significance of coronary ar-tery stenosis, patients undergo invasive coronary angiography (ICA). During ICA, fractional flow reserve (FFR), a quantitative marker of the functional significance of a stenosis \citep{Pijl96}, is determined. FFR is determined as the ratio of the invasively measured pressure under maximal hyperemic conditions, distal to a stenosis, relative to the pressure before the stenosis. The ideal FFR value is 1.0. FFR is currently considered the reference standard to determine the significance of coronary stenoses and is used to decide on the treatment \citep{Toni09}. Nevertheless, the FFR cut-off value that separates functionally significant from non-significant stenoses is not fully standardized: In clinical settings, cut-off values between 0.72 and 0.80 are commonly used \citep{Pijl96,De01,De08,Petr13b}, where stenoses with FFR measurements below the cut-off value are defined as functionally significant. Coronary CT angiography (CCTA) is often used to identify patients with suspected CAD as it allows noninvasive detection of coronary artery stenosis \citep{Budo08a}. Even though CCTA detects CAD with high sensitivity, it has limited specificity in determining the functional significance of the detected stenosis \citep{Meij08, Bamb11,Ko12}. Given the low specificity of CCTA, about 22-52\% of patients undergo invasive FFR measurements unnecessarily \citep{Ko12}. To reduce the number of unnecessary invasive procedures, establishing FFR noninvasively, i.e. directly from CCTA images, and identifying the functionally significant stenoses has become an area of intensive research. 

Thus far, the most successful and thoroughly evaluated noninvasive methods to determine the functional significance of coronary artery stenoses are based on the analysis of the blood flow through the coronary arteries. \citet{Tayl13} proposed a noninvasive CT-derived FFR ($FFR_{CT}$) measurement\review{, that was evaluated by \cite{Min12} and \cite{Norg14}}. This technology employs computational fluid dynamics applied to CCTA scans to determine FFR values in the coronary tree and thereby determines the functional significance of a stenosis. However, $FFR_{CT}$ requires accurate determination of the coronary artery tree geometry and the physiological boundary conditions. Hence, imaging artifacts such as blooming caused by large coronary calcifications, stents, and cardiac motion may compromise the accuracy of $FFR_{CT}$ \citep{Koo14}. 
\sout{$R2C1_4$}\review{ \cite{Itu12} presented a technique to estimate FFR in the coronary artery tree using a CCTA scan by simulating blood flow, which was later evaluated by \cite{,Renk14} and \cite{Coen15}. The method uses a patient-specific parametric lumped heart model, while modeling the hemodynamics in both healthy and stenotic vessel tree. Additionally, the method combines anatomical, hemodynamic and functional information from medical image as well as other clinical observation.}
\citet{Nick15} presented a technique to estimate FFR in the coronary artery tree using a CCTA scan by simulating blood flow using a patient-specific parametric lumped model. The authors modeled the coronary tree as an electrical circuit; volumetric flow rate was modeled as electrical current and pressure in the coronary artery as voltage. Using this analogy, the pressure in the coronary artery tree was simulated and FFR values were estimated within the coronary tree. While\sout{this technique} \review{these techniques \citep{Itu12,Nick15} }achieved high accuracy and real-time feedback,\sout{it} \review{they} strongly depend on the accuracy of coronary artery tree segmentation and its centerline determination, like the method described by \citet{Tayl13}. Manual delineation of the coronary artery centerline is a time consuming and cumbersome task, and most of the commercially available software packages occasionally require substantial manual interaction, especially in images with excessive atherosclerotic plaque or imaging artefacts \citep{Scha09a}. 

A number of approaches have been developed that do not rely on modeling the blood flow through the coronary arteries but exploit characteristics extracted from CCTA scans. For example, \citet{Stei10} presented a noninvasive approach to identify functionally significant coronary artery stenosis in a single CCTA scan using the transluminal attenuation gradient (TAG). TAG is defined as the gradient of the CT values attenuation along the artery lumen, and was shown to have a moderate positive correlation with the invasive FFR. However, like the method described by \citet{Nick15} and \citet{Tayl13}, computing TAG requires an accurate determination of coronary artery centerlines. 
Furthermore, \citet{Geor09} demonstrated that the comparison between myocardial regions imaged at rest and at stress may reveal myocardium perfusion defects that are caused by functionally significant coronary artery stenoses. This approach is interesting as it merges anatomical and functional information, obtained from CCTA scans. However, it requires acquisition of an additional CT scan which inevitably leads to a higher radiation dose, longer examination time, and injection of pharmacological stress agents. \citet{Xion15} presented a machine learning approach to classify patients with significant stenosis using myocardial characteristics derived from a single CCTA scan acquired at rest. The method automatically segments the left ventricle (LV) myocardium and aligns it with the standard 17-segments model \citep{Cerq02a}, which is used to relate a myocardial segment to its perfusing coronary artery. Supervised classification was employed to determine the significance of the evaluated stenosis based on \sout{R3C1}\review{ three hand-crafted features describing each myocardial segment.} \sout{ R3C1 the morphology and texture characteristics of the myocardial segments}Note that in this work stenosis was defined as significant based on the grade of the stenosis (with $\geq50\%$ luminal narrowing) and not by its effect on myocardial perfusion. Contrary to FFR, the grade of stenosis, \review{especially in the intermediate range 30\%-70\%}, is not necessarily related to its functional significance \citep{Toni09}. \sout{$R2C1_5$}\review{\cite{Han17} employed the method described by \cite{Xion15} to classify patients with functionally significant stenosis according to the invasively measured FFR.}

\begin{figure}
	\includegraphics[width=0.8\linewidth]{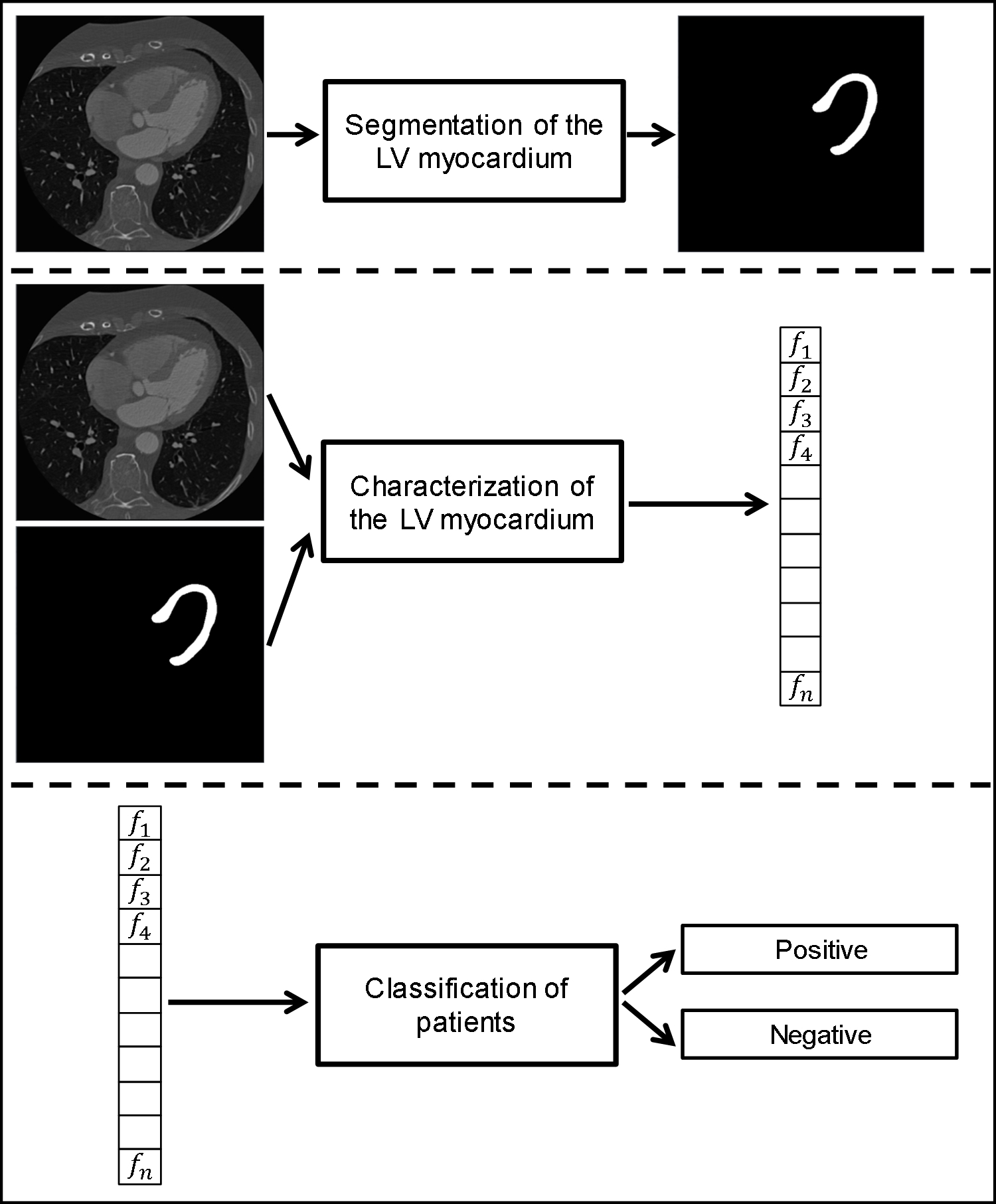}
	\caption{\review{Overview of the proposed algorithm. In a CCTA image, the LV myocardium is first segmented using a multiscale CNN. Then, from the segmented myocardium, encodings are extracted using a CAE and used to compute features([$f_1,f_2,f_3,..,f_n$]). These features are used to classify patients with functionally significant stenosis (positive) or without (negative) using SVM classification}}
	\label{fig:graphical_abstract}
\end{figure}

In this work, we present a novel method to \sout{automatically} identify patients with at least one functionally significant coronary artery stenosis in a single CCTA image acquired at rest. Given that obstruction of the blood flow in the coronaries may cause ischemia in the LV myocardium, only the myocardium is analyzed. This is in contrast to \sout{other}\sout{R2C10}\review{ most} methods that perform analysis of the coronary anatomy or stenosis and estimate the effect on the myocardium indirectly. In the proposed approach, %R4_minor
 \reviewminor{ deep learning, recently widely exploited in the analysis of medical images in a range of segmentation and detection tasks \citep{Litj17, Shen17,Zhou17}, is employed. First, }the LV myocardium is segmented using a multiscale convolutional neural network (CNN). Then, characteristics of the LV myocardium are extracted using convolutional autoencoder (CAE) \citep{Masc11,Beng13}. Subsequently, using the extracted characteristics, patients are classified with a support vector machine (SVM) classifier \citep{Cort95} into those with functionally significant stenosis and those without it. The reference for the functional significance of a coronary stenosis is provided by invasively determined FFR measurement, which is currently the clinical standard. \sout{$R2C6_1$}\review{The proposed approach is illustrated in Fig.~\ref{fig:graphical_abstract}.}

The remainder of the manuscript is organized as follows. Section \ref{data} describes the data and the reference standard. Section \ref{methods} describes the methods and Section \ref{eval} describes the evaluation procedure. Section \ref{results} reports our experimental results, which are then discussed in Section \ref{discussion}.

\section{Data} \label{data}

\subsection{Patient and Image Data}

This study includes retrospectively collected CCTA scans of 166 patients (age: $59.2 \pm 9.5$ years, 128 males) acquired between 2012 and 2016. The Institutional Ethical Review Board waived the need for informed consent. 

All CCTA scans were acquired using an ECG-triggered step and shoot protocol on a 256-detector row scanner (Philips Brilliance iCT, Philips Medical, Best, The Netherlands). A tube voltage of 120 kVp and tube current between 210 and 300 mAs were used. For patients $\le80$ kg contrast medium was injected using a flow rate of 6 mL/s for a total of 70 mL iopromide (Ultravist 300 mg I/mL, Bayer Healthcare, Berlin, Germany), followed by a 50 mL mixed contrast medium and saline (50:50) flush, and next a 30 mL saline flush. For patients $>80$ kg the flow rate was 6.7 mL/s and the volumes of the boluses were 80, 67 and 40 mL, respectively. Images were reconstructed to an in-plane resolution ranging from 0.38 to 0.56 mm, and 0.9 mm thick slices with 0.45 mm spacing.

\subsection{FFR Measurements}

\begin{figure}[]
	
\includegraphics[width=1.0\textwidth]{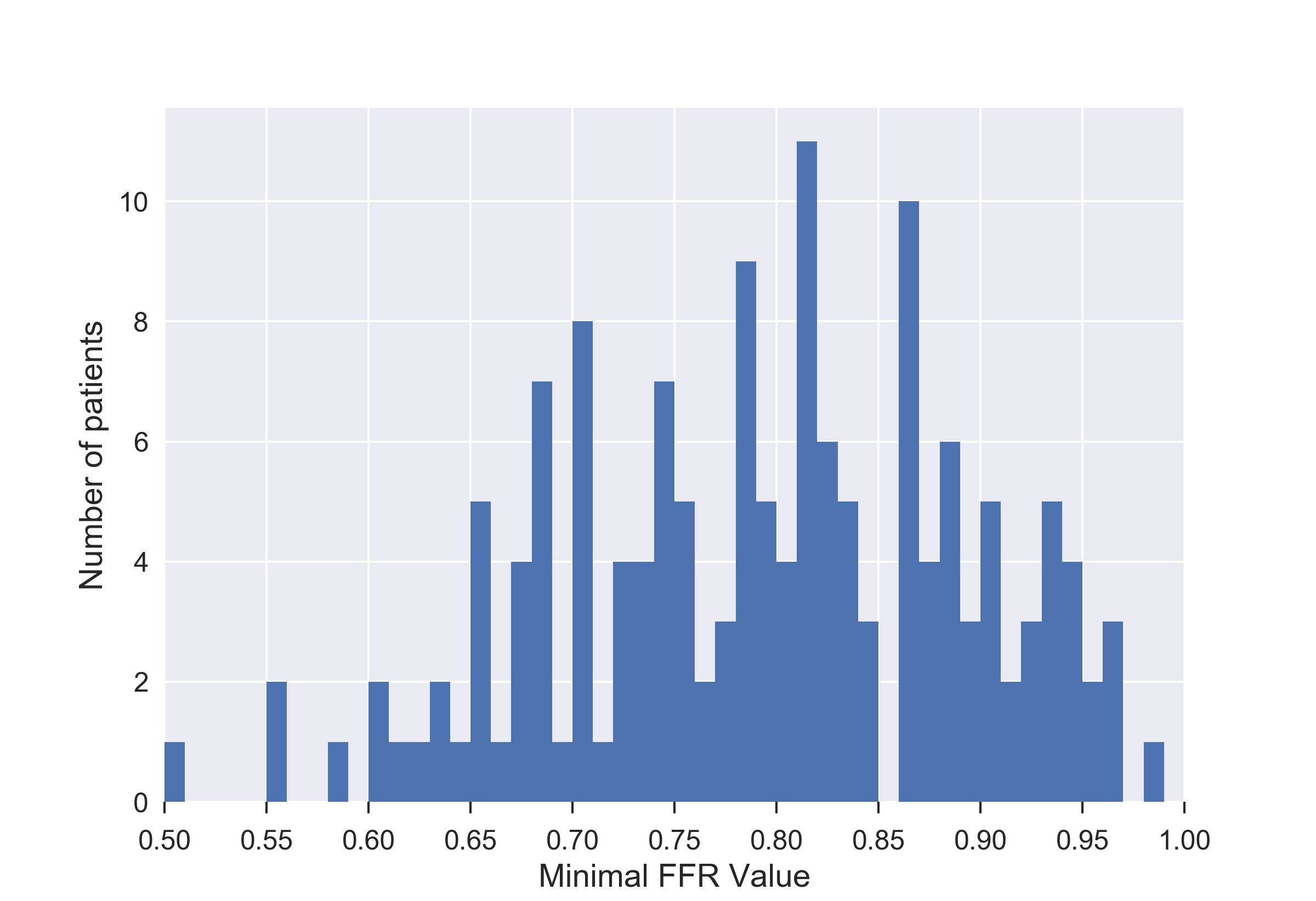}
		%  \vspace{2.0cm}

	\caption{A histogram of the minimal FFR values measured in 156 patients. }
	\label{fig:ffr_hist}
\end{figure}

Out of the 166 patients, 156 patients underwent invasive FFR measurements ($0.79 \pm 0.10$) within 1 year after the acquisition of the CCTA scan \sout{R3C3}\review{ (Median and interquartile range of time difference were 32.5 and 38.5 days, respectively). This cut-off value of 1 year was chosen because overall progression of coronary artery stenosis in patients with stable coronary artery disease is not expected \citep{Balk93}.} FFR was measured with a coronary pressure guidewire (Certus Pressure Wire, St. Jude Medical, St. Paul, Minnesota) at maximal hyperemia induced by intravenous adenosine, which was administered at a rate of 140 μg/kg per minute through a central vein. The FFR wire was placed as distally as possible in the target vessel and FFR was assessed by means of a manual pullback in the distal part of the target vessel (“standard FFR”). When multiple FFR measurements were available, or measurements for multiple stenoses were available, the minimum value was taken as the standard of reference for the patient. A histogram of the minimal invasively measured FFR values in the 156 patients is shown in Fig. \ref{fig:ffr_hist}.

\subsection{Manual Annotation of the LV Myocardium}

\begin{figure}[h]
	\centering
	\begin{subfigure}{0.5\textwidth}
		\centering
		\includegraphics[width=0.95\linewidth]{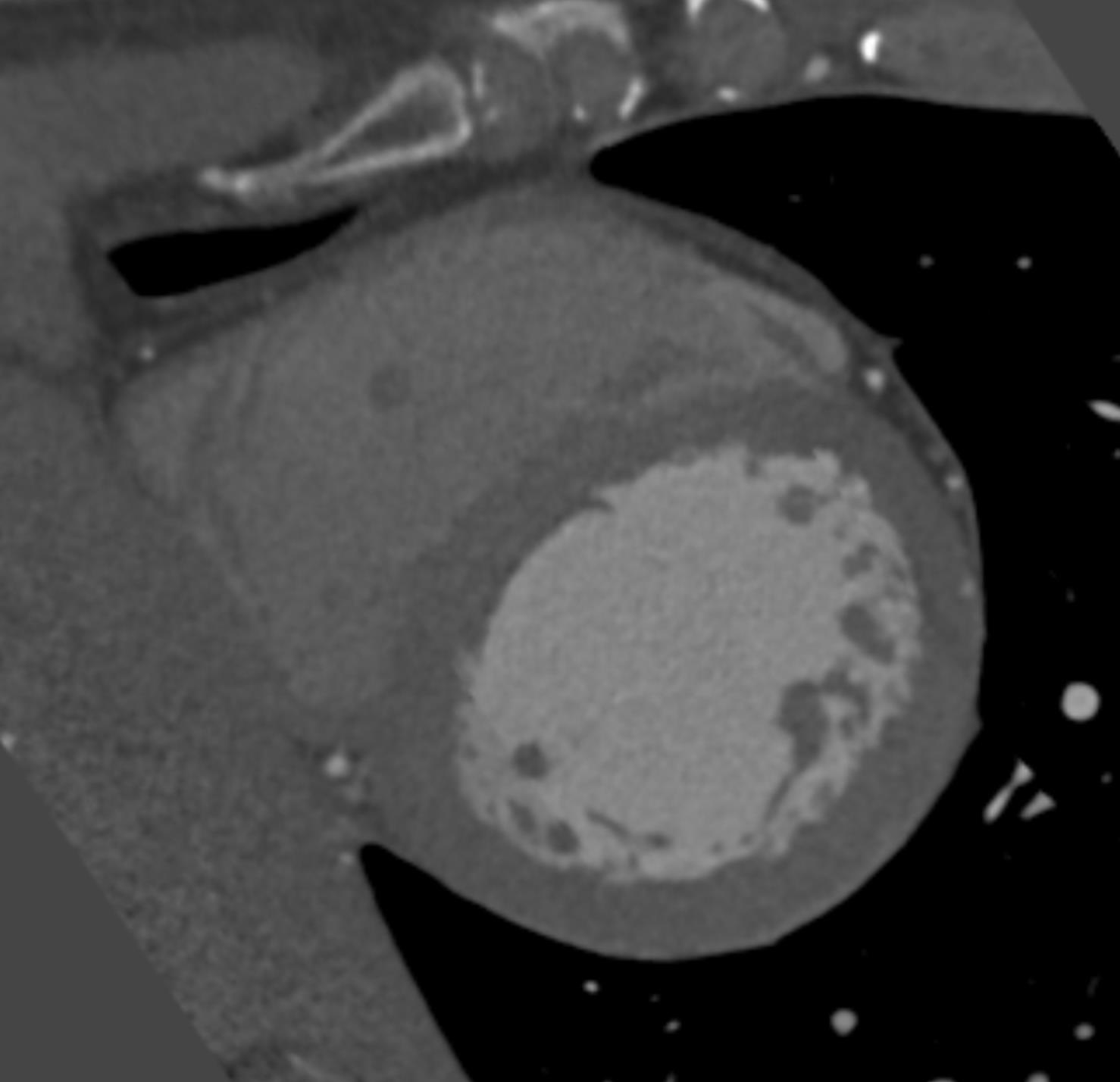}
		\caption{}
		
	\end{subfigure}%
	\begin{subfigure}{0.5\textwidth}
		\centering
		\includegraphics[width=0.95\linewidth]{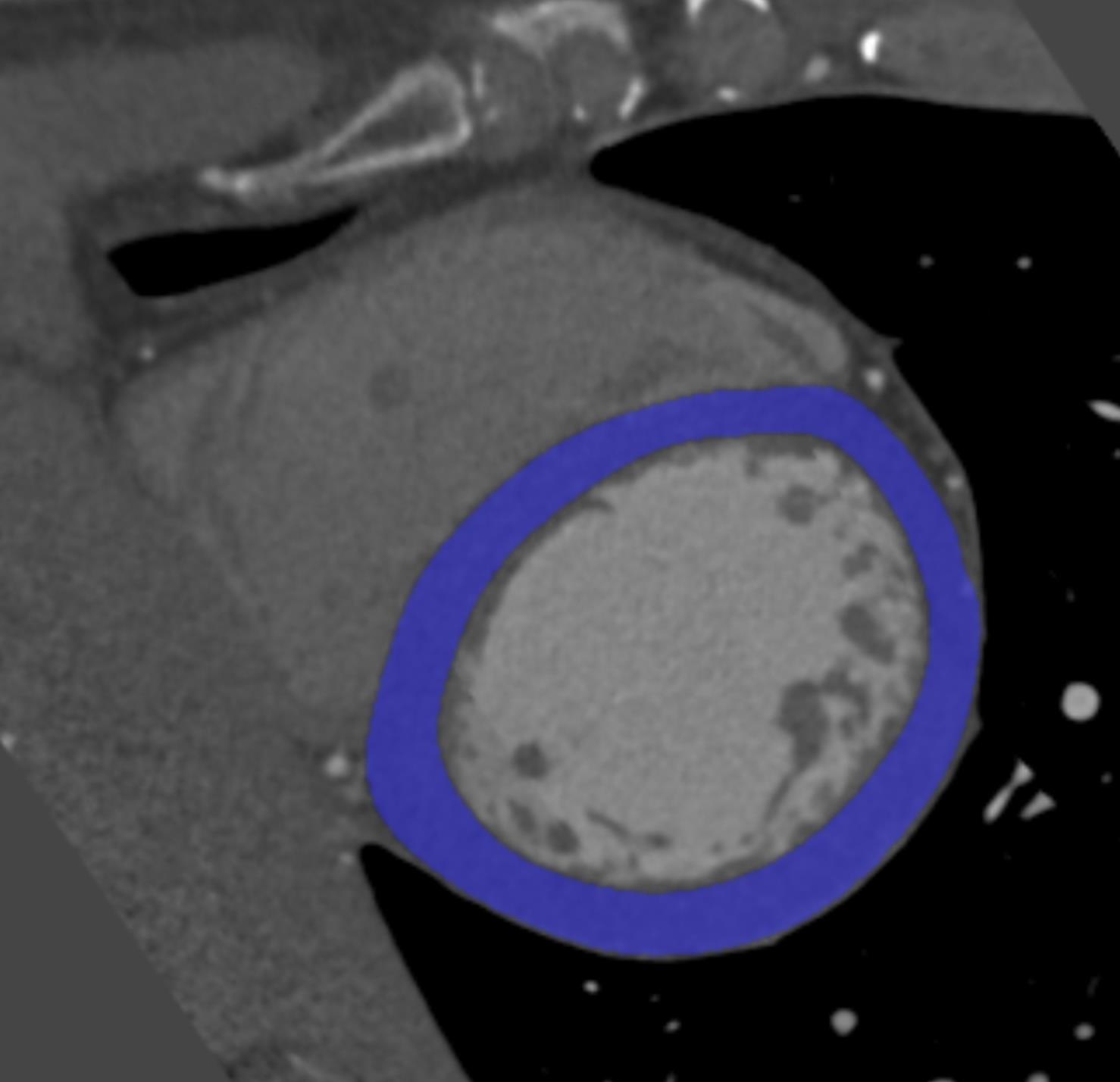}
		\caption{}
		
	\end{subfigure}
	\caption{(a) A short axis view showing LV myocardium in one slice of a CCTA and (b) the corresponding reference annotation of the LV myocardium. Only compact myocardium was segmented, while myocardial fat, papillary muscles and the trabeculae carneae were excluded.}
	\label{fig:obs1_obs2}
	
\end{figure}

To train, validate and test the segmentation of LV myocardium, the myocardium was manually annotated in 40 randomly selected CCTA scans. Manual annotations were performed by a trained observer using custom-built software created with the MeVisLab\footnote{http://www.mevislab.de} platform. Following clinical workflow, annotations were performed in the short axis view of the heart, while excluding myocardial fat, papillary muscles and the trabeculae carneae.
Segmentation was performed by manually placing points along the endocardium and epicardium in every third image slice. From the defined points, closed contours for the endocardium and the epicardium were created by cubic spline interpolation. The contours were propagated to the adjacent slices where they were manually adjusted when needed by moving existing or placing new points. Reference LV myocardium contained all voxels enclosed by the manually annotated endocardial and epicardial contours. Fig.~\ref{fig:obs1_obs2} illustrates a typical short axis view of the LV myocardium and the annotated manual reference.

\section{Methods} \label{methods}

To identify patients with functionally significant stenosis, the LV myocardium is first segmented using a multiscale CNN. Next, features are extracted from the segmented myocardium using a CAE, and these features are used to identify patients with functionally significant stenosis using an SVM classifier (Fig.~\ref{fig:graphical_abstract}).

\subsection{Myocardium Segmentation}

\sout{R1C1}\review{To date, several automatic methods for segmentation of the LV in CCTA have been proposed \citep{Tava13}. They can be divided in boundary-based (e.g. \cite{MP06,Zhen08a,Xion15}) and voxel-based segmentations (e.g. \cite{Kiri10}). The advantage of boundary-based approaches is their ability to perform subvoxel analysis. However, we aim to detect ischemic regions in the myocardium that normally exceed small subvoxel volumes \citep{Ross14}. Thus, we expect that voxel-based analysis offers sufficiently accurate segmentation for this task. Moreover, deep learning approaches, that perform voxel based analysis, have shown to outperform other methods in a number of segmentation tasks in medical images (e.g. \citet{Ronn15,Moes16,Hava17}), hence, we employ a CNN to automatically segment the LV myocardium. \sout{R4C2}To combine the analysis of local texture with distal spatial information, multiscale CNN is used}\sout{in this work, the LV myocardium is automatically segmented using a multiscale CNN} \citep{De15,Moes16,Kamn16,Hava17}. \sout{A preliminary version of the employed segmentation method has been presented in Zreik et al., 2016}

\begin{figure}[]

\includegraphics[width=1\linewidth]{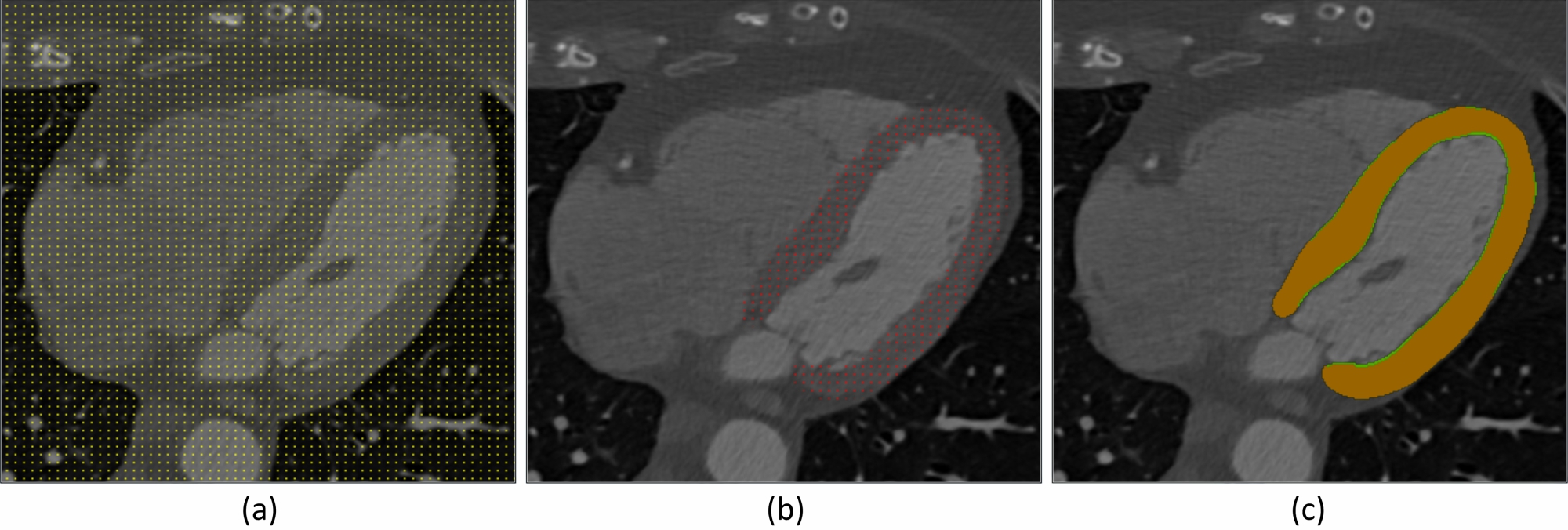}

\caption{ \review{Typical example of the LV myocardium segmentation stages. (a) Yellow dots show the equidistantly spaced voxels that are classified. (b) Red dots show the classification probability of the equidistantly spaced voxels. (c) \reviewminortwo{The union of} the green and \reviewminortwo{orange} masks shows the rough segmentation obtained by Gaussian blurring of the probabilities shown in (b), where its surface voxels (green mask) are iteratively classified. The orange mask shows the final segmentation result.}}

\label{fig:degrow_examples}

\end{figure}

The segmentation is performed in two stages. First, the LV myocardium is localized, and second, the voxels in the region of interest are classified. In our preliminary work \citep{Zrei16}, the LV myocardium was localized using a bounding box \citep{Vos17} created by an independent CNN. In this work, localization and voxel classification are both performed using the same CNN, circumventing the need for an independent localization method. 
Therefore, \sout{$R1C2_1$}\review{ to localize the LV myocardium}, equidistantly spaced voxels in the whole image are classified as LV myocardium or as background. By applying 3D Gaussian smoothing to the obtained sparse classification result followed by thresholding \review{of the smoothed probabilities,}\sout{an initial binary} \review{ a rough }segmentation of the LV myocardium is obtained. \sout{The obtained segmentation is likely inaccurate on its surface and accurate in the center. To further refine it, voxels located on the surface of the initial segmentation volume are classified iteratively until a stable result is obtained. A stable result is achieved when the labels of the surface voxels are not further altered by classification.}\review{Thereafter, to precisely segment the myocardium, only the voxels on the surface of the rough segmentation are iteratively classified as LV myocardium or as background. Note that reclassification of the voxels inside the rough segmentation is not needed as the myocardium is a compact structure. The iterative voxel classifications is repeated until all surface voxels are classified as LV myocardium. Fig.~\ref{fig:degrow_examples} illustrates the segmentation process.}

\begin{figure}[h!]

\centering
\subcaptionbox{Multiscale architecture\label{ms}}{%
	\includegraphics[width=1\textwidth]{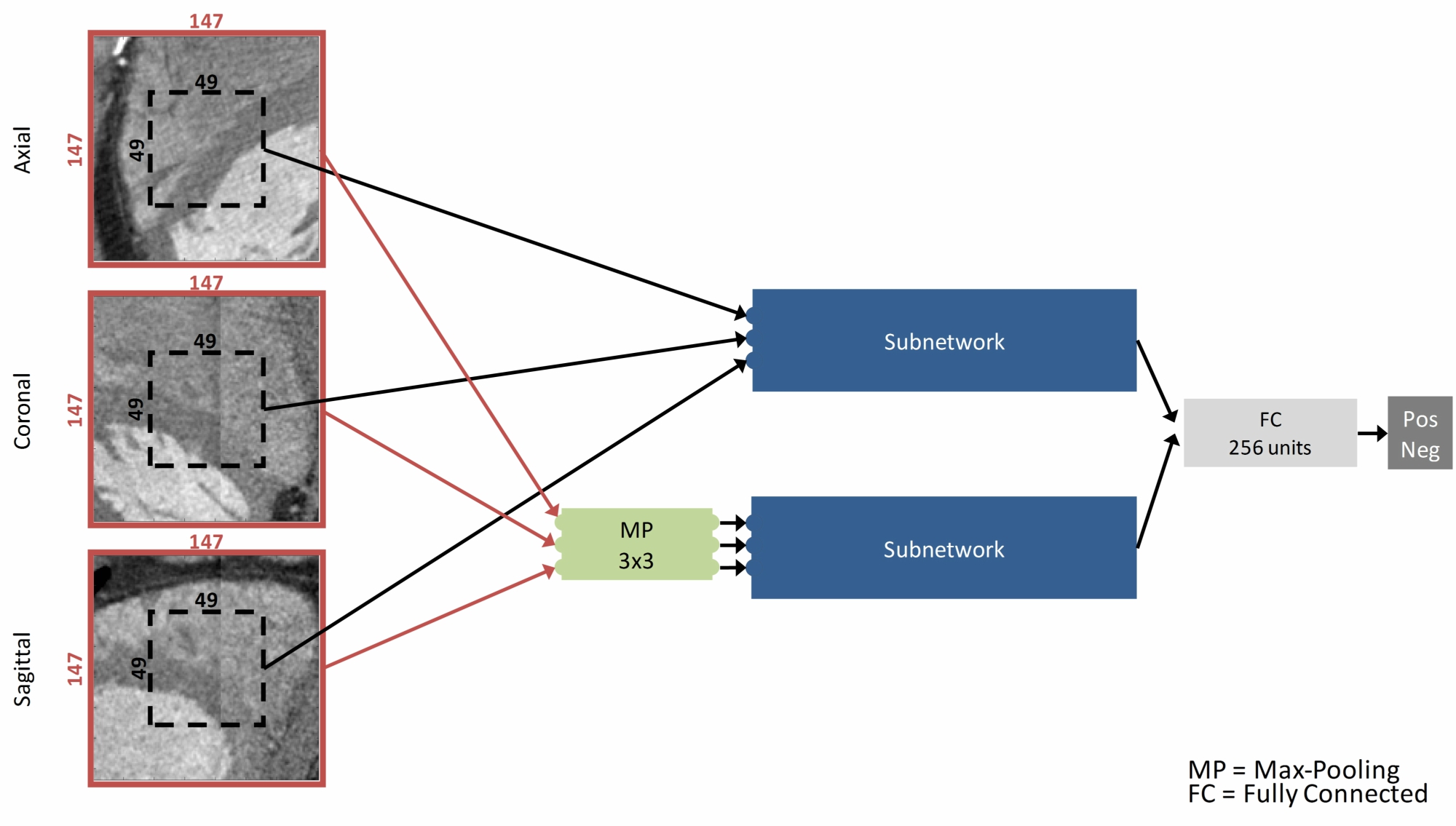}%
}\par\medskip
\subcaptionbox{SubNetwork\label{subnet}}{%
	\includegraphics[width=1\textwidth]{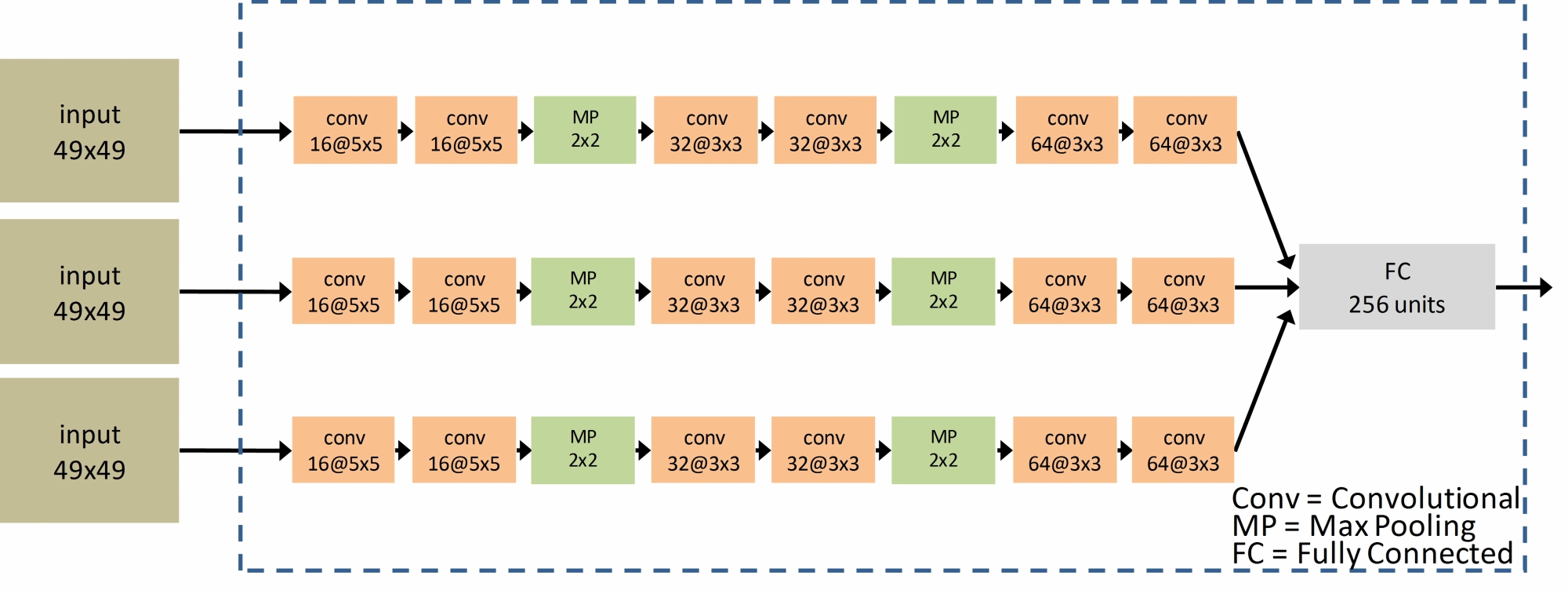}%
}
\caption{(a) Multiscale architecture: the multiscale CNN includes two identical subnetworks, each one analyzing a set of triplanar input patches taken at a single scale. The first set (black dotted squares) consists of three patches of $49\times 49$ voxels. The second set (red solid squares) consists of three patches of $147\times147$ voxels, that are downsampled to $49 \times 49$ voxels prior to CNN analysis (b) Subnetwork: Each subnetwork performs analysis of a set of orthogonal patches taken from axial, coronal and sagittal image slices with the target voxel in their centers.}
\label{fig:cnn_archticture}
\end{figure}

The CNN performs voxel classification using two sets of three orthogonal (2.5D) patches from axial, coronal and sagittal image slices with the target voxel in their centers (Fig.~\ref{fig:cnn_archticture}, (a)). The first set provides a small receptive field at high image resolution, and the second set provides a larger receptive field at lower image resolution. The multiscale approach enables the network to exploit both detailed local characteristics as well as contextual information. The first set consists of 3 patches of $49\times 49$ voxels and the second set of 3 patches of $147\times147$ voxels. The latter set of patches is downsampled by an additional $3\times3$ max-pooling layer resulting in patches of $49\times49$ voxels, as well.

To analyze both sets of patches, the CNN consists of two identical subnetworks (Fig.~\ref{fig:cnn_archticture}, (a)). Both subnetworks are fused together in a fully connected layer, followed by a softmax layer with two units providing a classification label of the voxel at hand. Each subnetwork (Fig.~\ref{fig:cnn_archticture}, (b)) consists of three streams that are fused together in a fully connected layer. \sout{R3C8}\review{Parameters of each stream layer are listed in Table \ref{layers}.} \sout{Each stream has two convolutional layers with 16 kernels of $5\times5$ each, which are followed by $2\times2$ max-pooling layer. The output of the first max-pooling layer is convolved in two convolutional layers with 32 kernels of $3\times3$ each, which are followed by a $2\times2$ max-pooling layer. The output is convolved again in two additional convolutional layers with 64 kernels of $3\times3$ each}In all fully connected layers, the number of units is set to 256. Batch normalization \citep{Ioff15} is used after each convolutional layer to make the training process faster and less sensitive to the learning rates. Exponential linear units (ELUs) \citep{clevert2015fast} are used in all layers as activation functions.

\subsection{Myocardial Characterization}\label{cae}

Functionally significant coronary artery stenosis causes ischemia in the LV myocardium and thereby\sout{likely} changes its texture characteristics in a %R2_minor1
\reviewminor{rest} CCTA image \citep{Niko06,Osaw16,Xion15,Han17}. In a CCTA image acquired at rest, where pharmacological stimulation is not applied to reveal reversible perfusion defects \reviewminor{(stress CCTA)} \citep{Geor09}, \review{the} texture changes \reviewminor{are less pronounced than in stress CCTA and} are mostly subtle \citep{Han17}. It would be extremely challenging to manually label myocardial voxels affected by ischemia. \sout{R3C4}\review{ Moreover, presently, no consensus has been established regarding the appearance of these changes in rest CCTA \citep{Spir13,Xion15,Han17}}. \reviewminortwo{Consequently, \sout{supervised} learning approaches for segmenting ischemic lesions would be hardly feasible. Therefore, in this work, the complete LV myocardium is described, omitting the need for segmentation of such lesions. This is performed by features that are learned in an unsupervised manner: by using a CAE}. The main purpose of a CAE, among the different unsupervised learning methods, is to extract general robust features from completely unlabeled data, while removing input redundancies and preserving essential aspects of the data in robust, compact and discriminative representations \citep{Masc11,Beng13,Lecu15}. In our work, a CAE is applied to myocardial voxels and their characteristics are used to identify patients with functionally significant stenosis. 

\begin{figure}[h]
	
	\begin{minipage}[b]{1.0\textwidth}
		\centering
		\centerline{\includegraphics[width=1.0\textwidth]{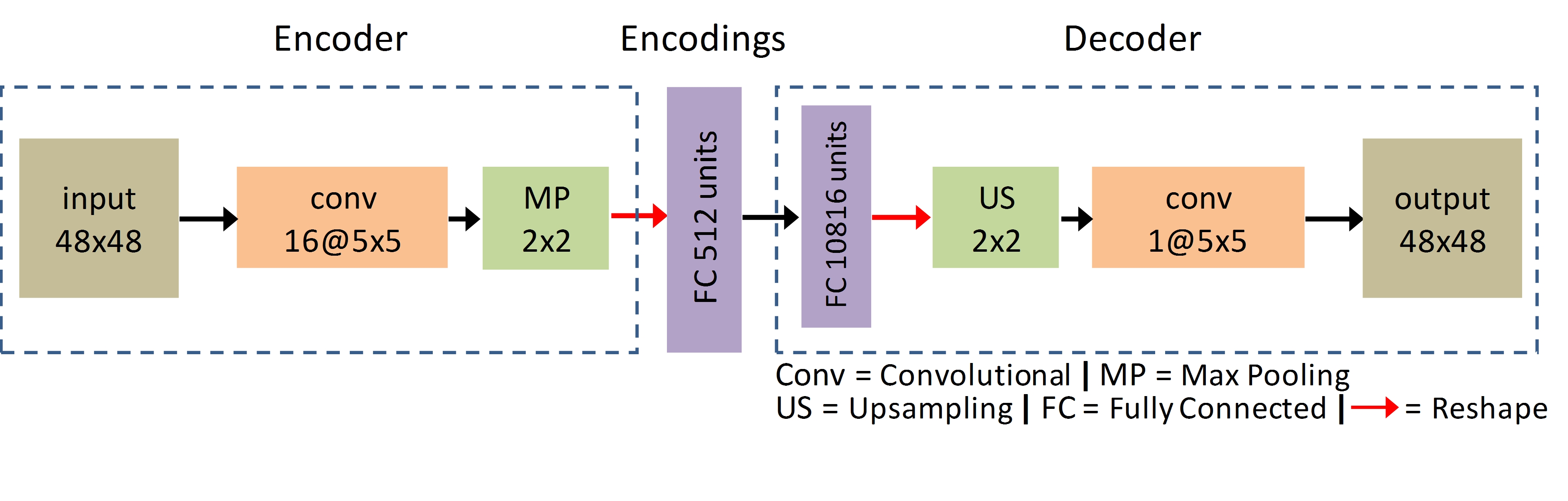}}
		%  \vspace{2.0cm}
		
	\end{minipage}
	\caption{CAE architecture: The input for the CAE is an axial patch around a myocardial voxel. The encoder consist of one convolution layer followed by max-pooling and a fully connected layer with 512 units. The decoder consists of one fully connected layer, one upsampling layer followed by a convolution layer providing the reconstructed input. }
	\label{fig:cae}
\end{figure}

A typical CAE contains two major parts, an encoder and a decoder \citep{Masc11,Beng13}. The encoder compresses the data to a lower dimensional representation by convolutional operations and max-pooling. The decoder expands the compressed form to reconstruct the input data by deconvolutional operations and upsampling. The coupling between the encoder and the decoder, and minimizing the loss between the input and the output ensure that abstract features (encodings) generated from the input contain sufficient information to reconstruct it \citep{Beng13}. 
The CAE architecture used in this work is shown in Fig.~\ref{fig:cae}. The input of the CAE comprises of $48\times48$ voxels axial patches around myocardial voxels.\sout{R3C8} \review{ The CAE parameters are listed in Table \ref{layers}.} \sout{ The encoder consists of one convolutional layer with 16 kernels of $5\times5$ and one $2\times 2$ max-pooling layer compressing the input, and one fully connected layer with 512 units providing the encodings. The decoder consists of one fully connected layer with 10,816 units, one $2\times 2$ upsampling layer followed by one convolution layer with a single $5\times5$ kernel}The output of the decoder is the reconstructed input patch.
Batch normalization is used after each convolutional layer to make the training process faster and less sensitive to learning rates. ELUs are used as activation functions in all layers except the output layer, where nonlinearity is not applied. Once the CAE is trained, the decoder is removed and the fully connected layer becomes the output layer which is used to generate encodings for unseen patches.

% \usepackage{graphicx}
\begin{table}[]
	\centering
	\caption{\review{Parameters of the one stream in subnetwork (Fig.~\ref{subnet}) of the multiscale CNN used for segmentation of the LV myocardium (left) and the CAE (Fig.~\ref{fig:cae}) used to encode the LV myocardium (right). Types of the networks layers, filter sizes and activation functions are listed. Abbreviations: Conv = convolutional, MP = max pooling, FC = fully connected, ELU = exponential linear unit, and US = upsampling.}}
	\label{layers}
	\resizebox{\textwidth}{!}{%
		\begin{tabular}{lcc|lcc}
			\hline
			\textbf{\begin{tabular}[l]{@{}l@{}}Layers of one \\ stream in subnetwork\end{tabular}} & \textbf{Size}           & \textbf{Activation} & \textbf{\begin{tabular}[l]{@{}l@{}}Layers of \\ CAE\end{tabular}}           & \textbf{Size}           & \textbf{Activation}  \\ \hline
			Input Layer                                                                            & $1 \times 49 \times 49$ & -                & Input Layer                                                                 & $1 \times 48 \times 48$ & -                 \\
			Conv Layer                                                                             & $16 \times 5 \times 5$  & ELU                 & Conv Layer                                                                  & $16 \times 5 \times 5$  & ELU                  \\
			Conv Layer                                                                             & $16 \times 5 \times 5$  & ELU                 & MP Layer                                                              & $ 2 \times 2$           & -                 \\
			MP Layer                                                                         & $ 2 \times 2$           & -                & \begin{tabular}[c]{@{}c@{}}FC Layer\end{tabular} & 512                     & ELU                  \\
			Conv Layer                                                                             & $32\times 3 \times 3$   & ELU                 & FC Layer                                                       & 10,816                  & ELU                  \\
			Conv Layer                                                                             & $32\times 3 \times 3$   & ELU                 & US Layer                                                             & $ 2 \times 2$           & -                 \\
			MP Layer                                                                         & $ 2 \times 2$           & -                & Conv Layer                                                                  & $1 \times 5 \times 5$   & -                 \\
			Conv Layer                                                                             & $64\times 3 \times 3$   & ELU                 & Output Layer                                                                & $1 \times 48 \times 48$ & -                 \\
			Conv Layer                                                                             & $64\times 3 \times 3$   & ELU                 & \multicolumn{1}{l}{}                                                        & \multicolumn{1}{l}{}    & \multicolumn{1}{l}{}
		\end{tabular}%
	}
\end{table}

\subsection{Patient Classification}

Patients with at least one significant coronary artery stenosis are identified based on the characteristics of the LV myocardium that are encoded by the CAE. 
The reference for the presence of functionally significant coronary artery stenosis is provided by the invasive FFR measurements.

\begin{figure}[h!]

		\includegraphics[width=0.8\textwidth]{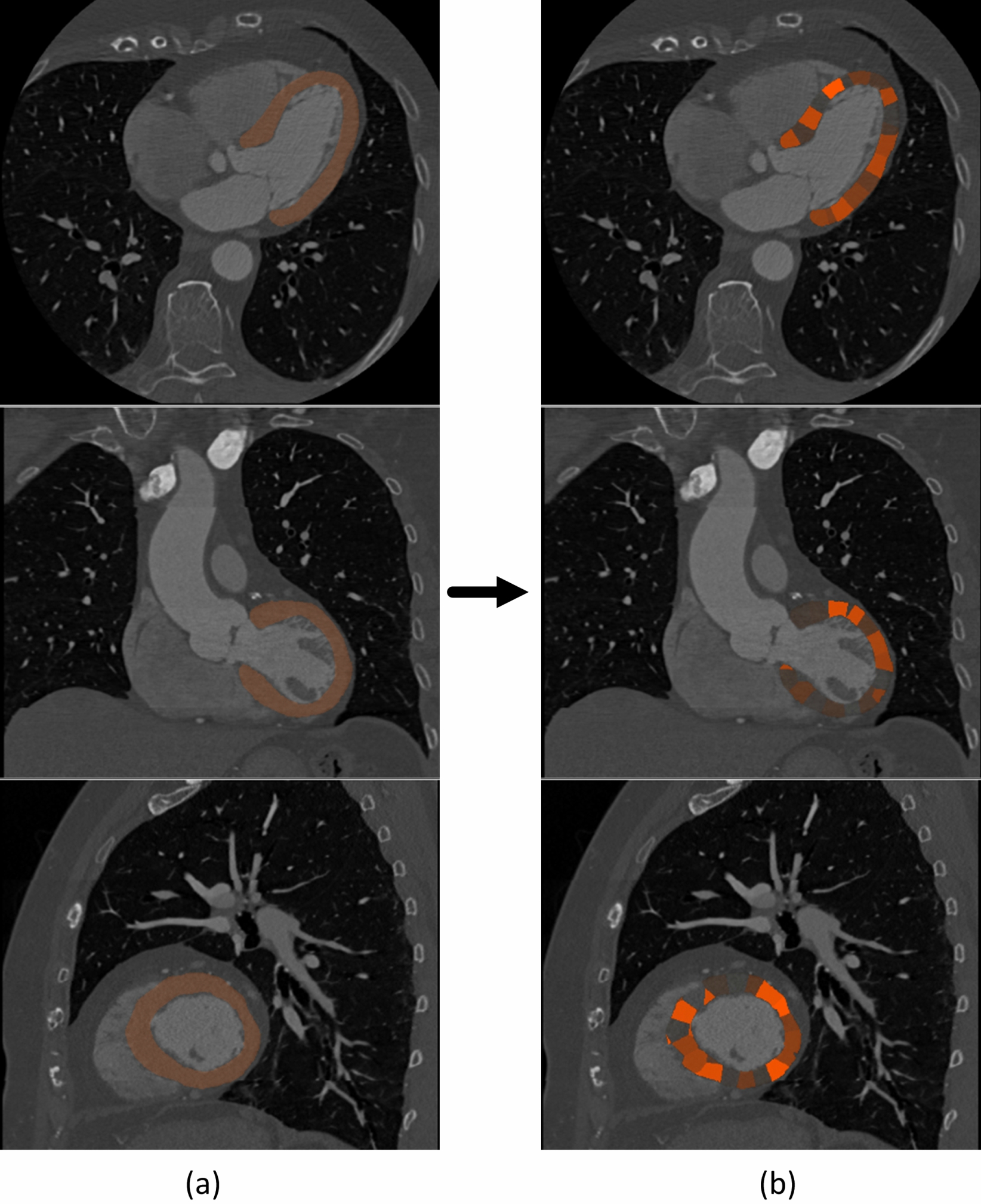}
		%  \vspace{2.0cm}

	\caption{\review{(a) An example of axial (top), sagittal (middle) and coronal (bottom) image slices of a CCTA with segmented LV myocardium voxels. (b) The segmented LV myocardium voxels are clustered using K-means algorithm. Different colors represent different clusters.}}
	\label{fig:clusters}
\end{figure}

Since a functionally significant stenosis is expected to have a local impact on the myocardial blood perfusion\sout{, and consequently on the texture characteristics of the contrast enhancement of the hypoperfused regions} \citep{Mehr11,Ross14}, the LV myocardium is divided into 500 spatially connected clusters. Clustering is achieved using the fast K-means algorithm \citep{Scul10}, based on the spatial location of the myocardial voxels. \sout{$R1C3_1$}\review{ A typical example of such a clustering is shown in Fig.~\ref{fig:clusters}.} 

Within a single cluster, a large variance in an encoding likely indicates an aspect of the cluster's inhomogeneity, and thereby the presence of abnormal myocardial tissue. \sout{$R1C4_1$}In preliminary experiments, a number of statistics of the encodings was evaluated. \reviewminortwo{The best result was obtained using the standard deviation of each of the encodings over all voxels within a cluster. 
\sout{To describe the whole LV myocardium rather than its clusters, the maxima of the standard deviations of each encoding over all clusters are used and inspired by the multi instance learning approach, namely the statistics kernel approaches as proposed by \cite{Gart02}, we determined the maxima of the standard deviations to use as features describing each patient.}%R2minor_3 R4M2
Inspired by the multi instance learning approach \citep{Gart02}, we have described the whole LV myocardium, rather than its clusters. Therefore, for each encoding the maximum of all standard deviations over the clusters is determined. 
This results in a vector of features describing each patient.}

 Finally, based on the extracted and clustered features, patients are classified into those with a functionally significant coronary artery stenosis or those without it.\sout{R1C5} \review{ As training an end-to-end deep learning system for patient classification is not feasible due to the small size of the available dataset, patient classification is performed} using an SVM classifier with a radial basis function.

\section{Evaluation}\label{eval}

\subsection{Myocardium Segmentation}\label{myoeval}

Manual annotation of the LV myocardium is a time-consuming task. Therefore, automatic segmentation was evaluated quantitatively in a subset of test scans with manually defined reference and qualitatively in all test scans. 

Quantitative evaluation was performed using the Dice coefficient, as an overlap measure between reference and automatically segmented volumes. In addition, the mean absolute surface distance (MAD) between the reference and automatically segmented LV boundaries was computed. 

Qualitative evaluation was performed by an expert who visually inspected and graded the automatic segmentation using the quality grades as defined by \citet{Abad10}. 

\subsection{Patient Classification}
Classification of patients into those having functionally significant stenosis or those without it was evaluated using a receiver operating characteristic (ROC) curve. %Patients in whom qualitative evaluation of the myocardium segmentation is graded 4 or 5, i.e. those cases where segmentation is highly inaccurate or failed, are excluded from further analysis.

\section{Experiments and Results}\label{results}

\subsection{Myocardium Segmentation}\label{experiments_seg}

To train the multiscale CNN for LV myocardium segmentation, 20 manually annotated scans were randomly selected (training set) from the set of 40 scans with manual annotations. The remaining 20 scans were used as an independent test set and used for quantitative evaluation of the segmentation (see Section \ref{myoeval}). All CNN hyperparameters and optimal thresholds were determined in preliminary experiments using the training set.

Two sets of triplanar patches were extracted around positive (LV myocardium) and negative (background) voxels with the voxel of interest in their centers (Fig.~\ref{fig:cnn_archticture}, (a)). Mini-batches of image patches, extracted from the training set and balanced with respect to class labels, were used for training the CNN. Given that many background voxels distant to the LV myocardium, such as those representing lungs or bones, do not resemble myocardium while some voxels proximal to it are almost indistinguishable from it,\sout{$R4C3_1$} \review{ and the LV myocardium comprises only a small part of the image,} training voxels were selected according to their distance to the LV myocardium. In particular, during training, background voxels in the vicinity of the myocardium ($< 80$ voxels) were selected four times more often than those further away from it.\sout{$R4C3_2$} \review{ These parameters were chosen empirically using the training set.} \sout{$R4C3_3$}The training process \sout{was performed over}\review{converged after} 200 epochs, \review{where} each epoch consisted of 200 mini-batches, and each mini-batch consisted of 500 patches selected from the training data. A random dropout \citep{Sriv14} of 50\% was applied during training in each fully connected layer to prevent overfitting. Stochastic gradient descent with Nesterov momentum and learning rate $0.1$ was used to minimize the cross-entropy loss function \citep{Nest07}.

\sout{$R1C2_2$}During testing, to localize the LV myocardium, equidistantly spaced voxels were classified (every $5^{th}$ voxel) in the axial, coronal and sagittal image planes. %(Fig. \ref{fig:flow_segmenation}). 
The obtained classification result was smoothed using a 3D Gaussian filter with a kernel size of 5 voxels and thereafter thresholded at \sout{posterior probability of}0.5 to obtain an initial \sout{binary segmentation} \review{rough segmentation}. The surface voxels of the initial \review{rough} segmentation were iteratively classified until the final segmentation was obtained.

\begin{figure}[]%
		{\includegraphics[width=0.8\linewidth]{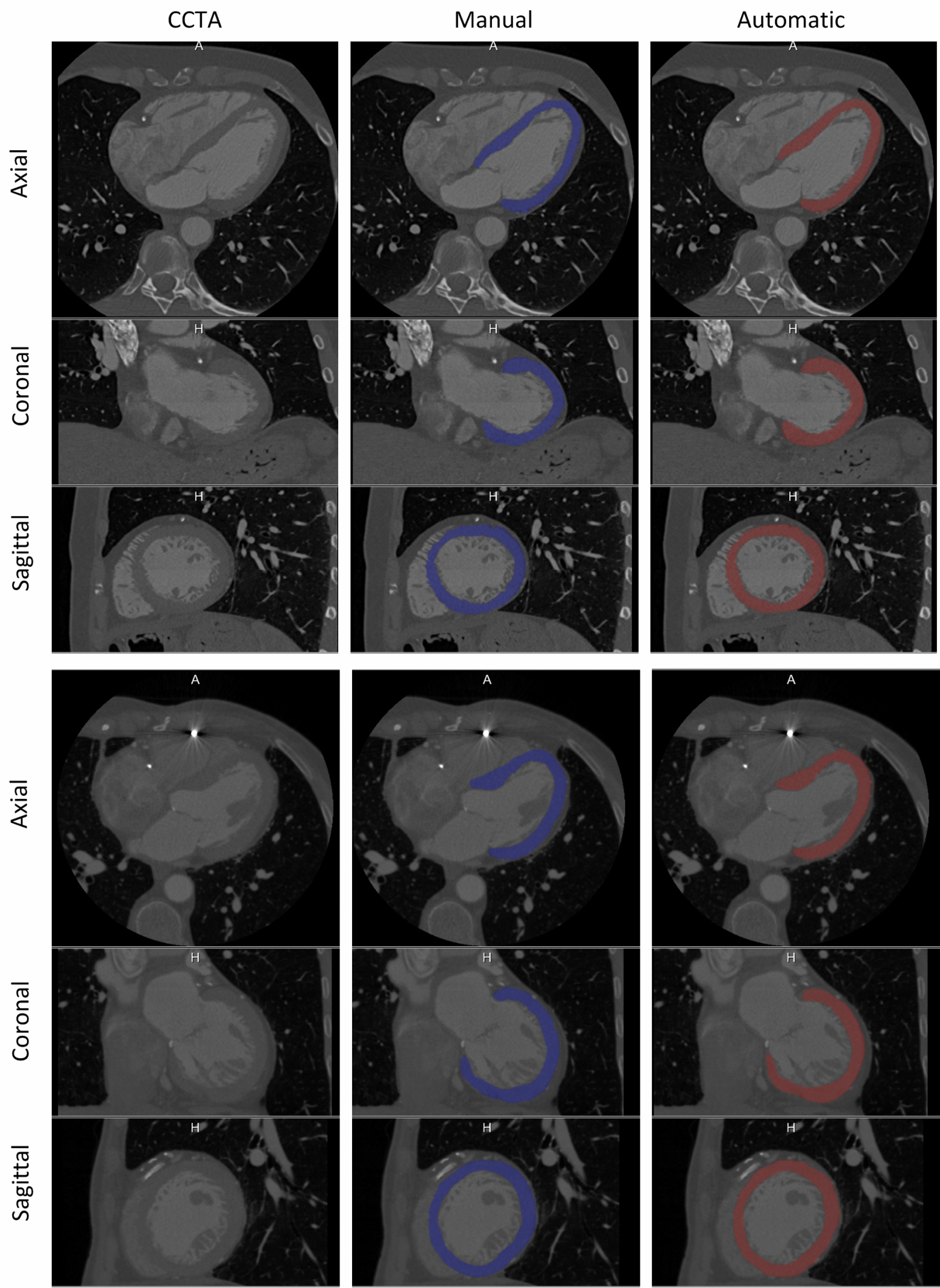}}
		%  \vspace{2.0cm}
	\caption{Slices from a CCTA scan, with manual annotation (blue) and automatic segmentation (red) are shown for two randomly selected test scans. For each scan an axial (top), coronal (middle), and sagittal (bottom) slices are shown.}
	\label{fig:examples_3_auto_with_manual}
\end{figure}

Fig.~\ref{fig:examples_3_auto_with_manual} illustrates segmentation results. Quantitative evaluation of the segmentation performed on the 20 test scans resulted in a Dice coefficient of $91.4 \pm 2.1 \%$ and a MAD of $0.7 \pm 0.1$ mm. %Fig.~\ref{fig:degrow_examples} and
Qualitative evaluation of the segmentation was performed in all 146 test scans. Results are summarized in Table \ref{list:quality_results}. \sout{and examples of automatic segmentations assigned different grades are shown in Fig.~\ref{fig:grade_examples}.}

\begin{table}[h!]
	\centering

	\begin{tabular}{lc}
			\toprule
		Grade              &  Result (Scans/Total scans) \\
	\hline
		
		1 - Very accurate            	 &  $~~~~74.0\%$ ($108/146$)   \\
		
		2 - Accurate            		 &  $~~15.8\%$ ($23/146$)  \\    
		
		3 - Mostly accurate             	& $3.4\%$ ($5/146$)  \\    
		
		4 - Inaccurate            			 & $6.1\%$ ($9/146$)  \\    
		
		5 - Segmentation failed            & $0.7\%$ ($1/146$)  \\
		
	\end{tabular}
	\caption{ Qualitative results of the automatic segmentation of LV myocardium segmentation in 146 scans. Segmentations were visually examined and qualitative grades were assigned as defined by \cite{Abad10}. Table lists percentage of scans (Result) assigned each grade (Grade).}

	\label{list:quality_results}	
	
\end{table}

\subsection{Myocardial Characterization}\label{experiments_cae}

The CAE was trained and validated using 20 training scans with the corresponding manual annotations of the LV myocardium. Axial patches of $48\times48$ voxels around randomly selected voxels of the myocardium were extracted from these scans. 
$90\%$ of the extracted patches were randomly selected for training the CAE and the remaining $10\%$ for validating it. Training was performed over 750 epochs, each epoch consisted of 200 training and 20 validation mini-batches. Each mini-batch consisted of 500 patches, which were randomly selected from training and validation data. Stochastic gradient descent with Nesterov momentum and learning rate $10^{-5}$ was used to minimize the loss function which was defined as mean squared error between the input patch and the reconstructed patch. 

\begin{figure}[h!]
	
	\begin{minipage}[b]{1.0\linewidth}
		\centering
		\centerline{\includegraphics[width=1\linewidth]{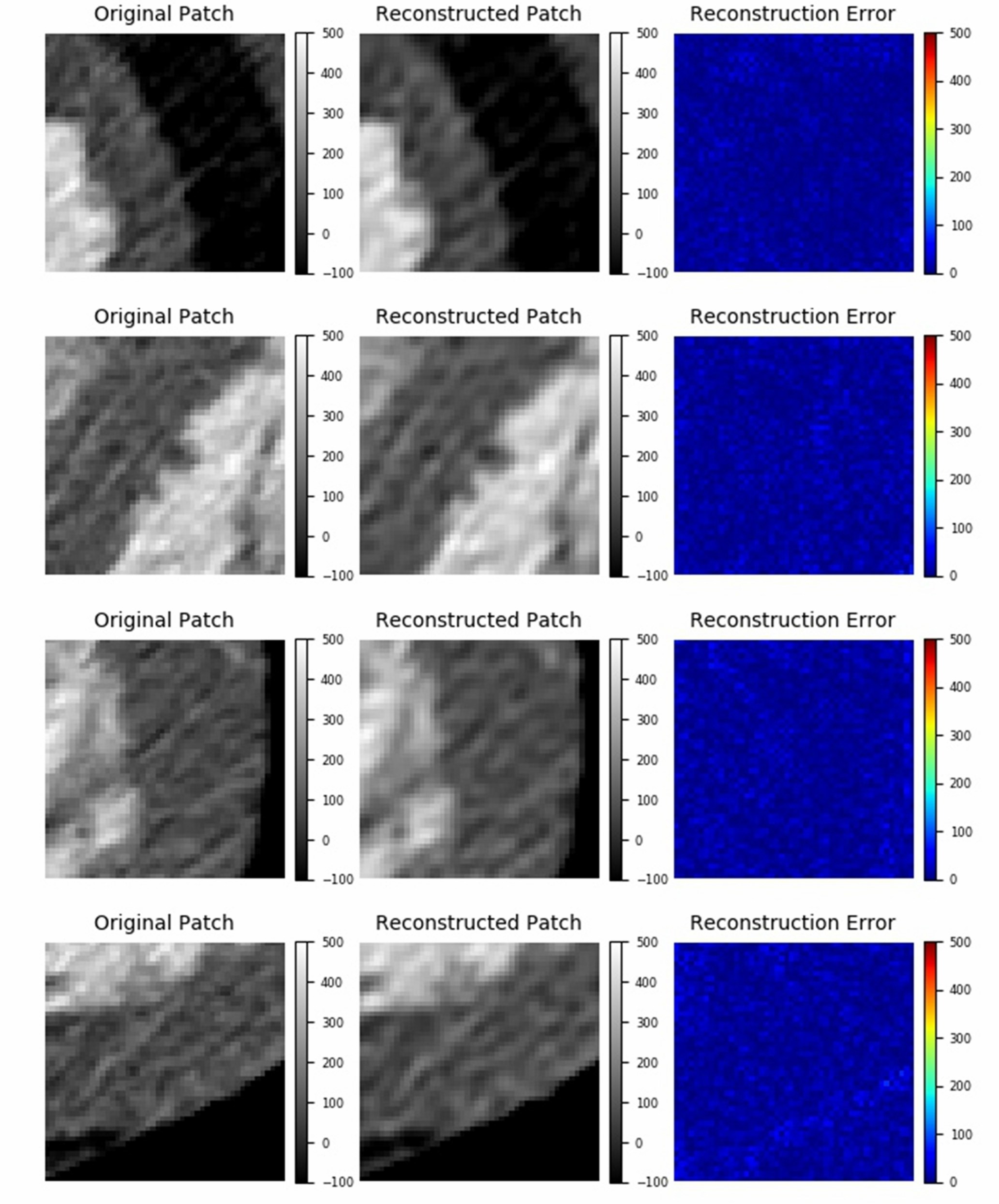}}
		%  \vspace{2.0cm}
		
	\end{minipage}
	\caption{Four examples of reconstructed patches by CAE randomly selected from test scans. Each row contains the original $48\times48$ input patch (left), reconstructed $48\times48$ output patch (middle)\review{, and the reconstruction error (right).}}
	\label{fig:reconstructed_patches}
\end{figure}

\sout{Fig.~\ref{fig:cae_loss} shows the average training and validation loss per epoch.} Fig.~\ref{fig:reconstructed_patches} illustrates pairs of input patches and the corresponding reconstructed patches \sout{R4C10}\review{ and reconstruction errors}, randomly selected from the LV myocardium in test scans.

\subsection{Patient Classification}

Since the aim of this study was to examine the feasibility of using only information derived from the LV myocardium to identify patients with significant coronary artery stenosis, only patients with accurate automatic segmentation that includes \sout{nearly} complete LV myocardium (grades 1, 2 or 3) were included in the analysis.\sout{R3C15} Moreover, \review{CCTA images which} \sout{patients whose scans}were used for training the LV myocardium segmentation or characterization were excluded from further analysis to completely separate training and test data. This resulted in a set of 126 patients out of the initial 166 patients set. 

Since invasive FFR measurements provided a reference for classification of patients according to the presence or absence of functionally significant stenosis, a cut-off value on the FFR measurements had to be defined. In the literature, different FFR cut-off values, ranging from 0.72 to 0.8, have been used \citep{Pijl96,De01,Sama06,De08,Petr13b}. In this study the cut-off value was set to 0.78. This value provided the best balance between positive and negative class, thus patients with a minimum FFR measurement below or equal to 0.78 were considered positive (64 patients), and those with a minimum FFR above 0.78 were considered negative (62 patients). 

\begin{figure}[h]
	
	\begin{minipage}[b]{1.0\linewidth}
		\centering
		\centerline{\includegraphics[width=\linewidth]{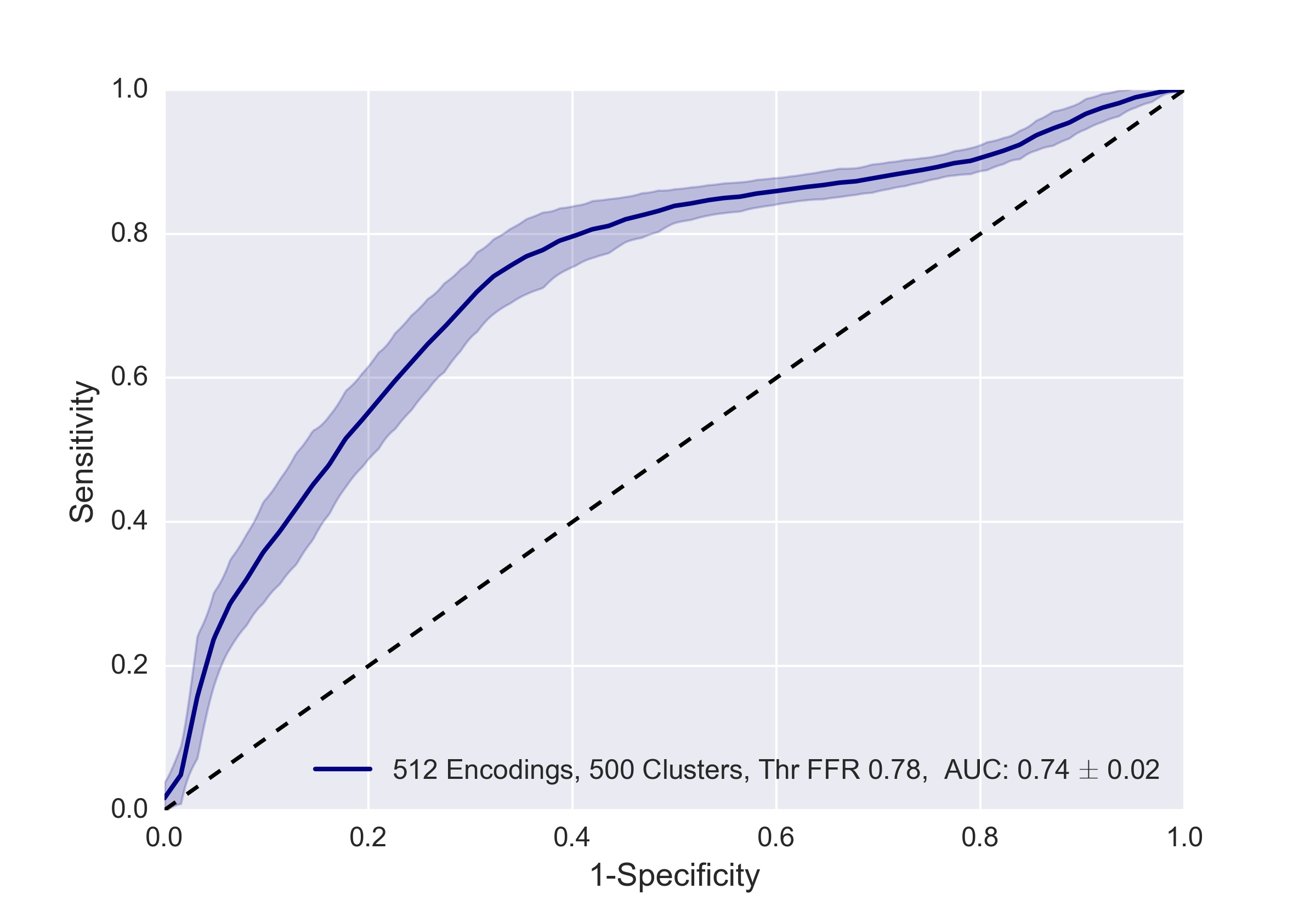}}
		%  \vspace{2.0cm}
		
	\end{minipage}
	\caption{Average ROC curve for classification of patients using 500 clusters and 512 encodings of the LV myocardium and the cut-off value on FFR set to 0.78. The shaded area represent the standard deviation of the sensitivity.}
	\label{fig:loo_roc}
\end{figure}

Classification was performed using characteristics of the segmented LV myocardium, where its voxels were encoded by 512 encodings and subsequently, features were extracted from its 500 clusters. Evaluation of patient classification was performed in 10-fold cross-validation experiments. To evaluate the robustness of the method, the cross-validation experiments were performed 50 times with re-randomized folds. Optimal SVM parameters (C and $ \gamma $) were selected in every experiment using a grid search on the training set only. 
The results are shown in Fig.~\ref{fig:loo_roc}. An average AUC of $0.74\pm 0.02$ was achieved, while for the sensitivities set to 0.60, 0.70 and 0.80, corresponding specificities were 0.77, 0.71 and 0.59, respectively.

\subsection{The Effect of Number of Clusters}\label{clustersresults}

To evaluate the effect of the number of used clusters on patient classification, additional experiments were performed where the LV myocardium was clustered into different number of clusters, namely 1, 10, 20, and 1000. In these experiments, the number of encodings was set to 512. The results are illustrated in Fig.~\ref{fig:perf_clusters_encodings_sizes}, (a). The \review{obtained} AUC ranged \sout{for different numbers of clusters} from 0.62 to 0.74, while the highest AUC was obtained with 500 clusters.

\subsection{The Effect of Number of Encodings}

\begin{figure}[h!]
	
	\begin{subfigure}{0.45\textwidth}
\centering
		\centerline{\includegraphics[width=1.1\linewidth]{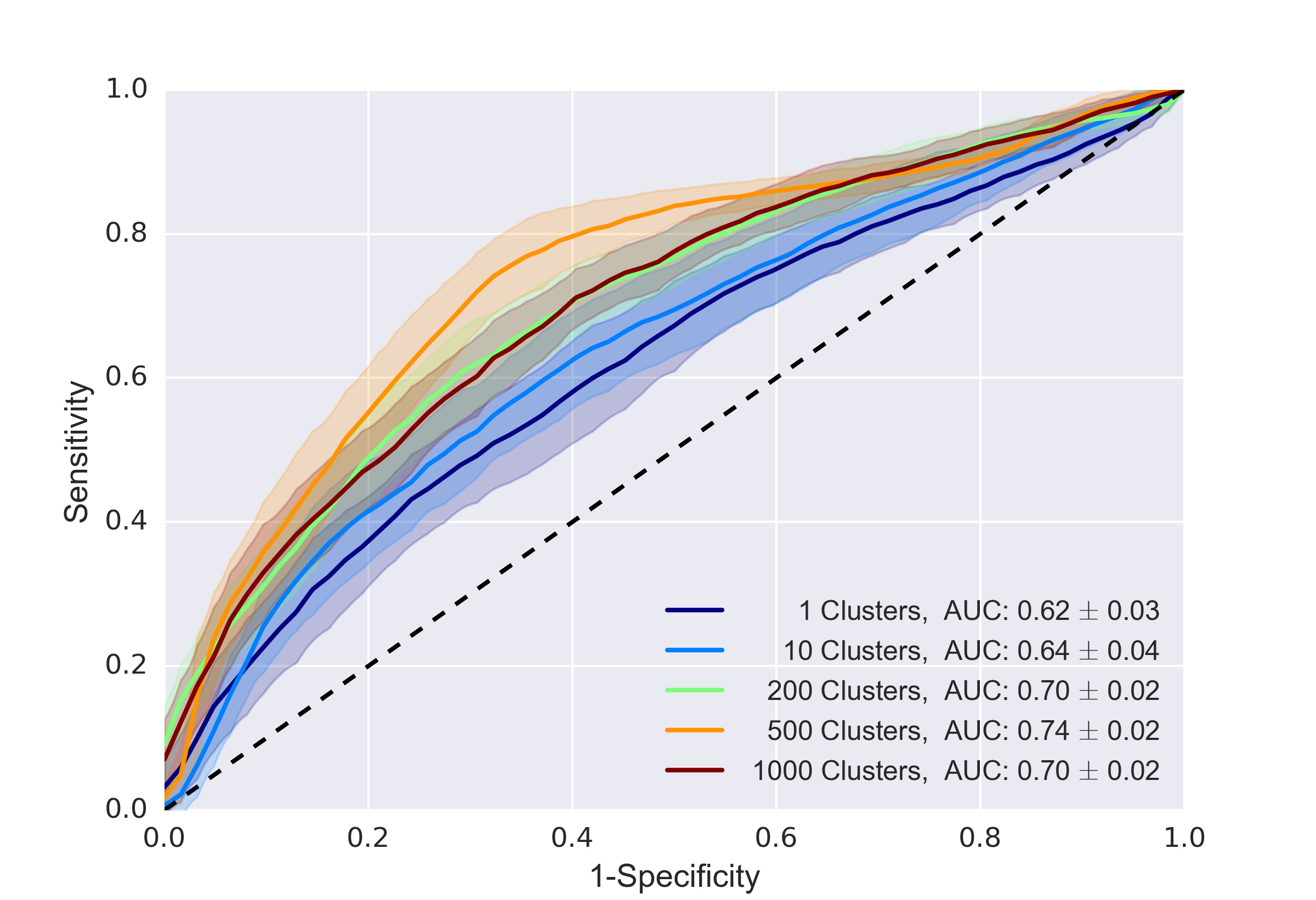}}
		\caption{}
		
	\end{subfigure}
	\begin{subfigure}{0.45\textwidth}
\centering
		\centerline{\includegraphics[width=1.1\linewidth]{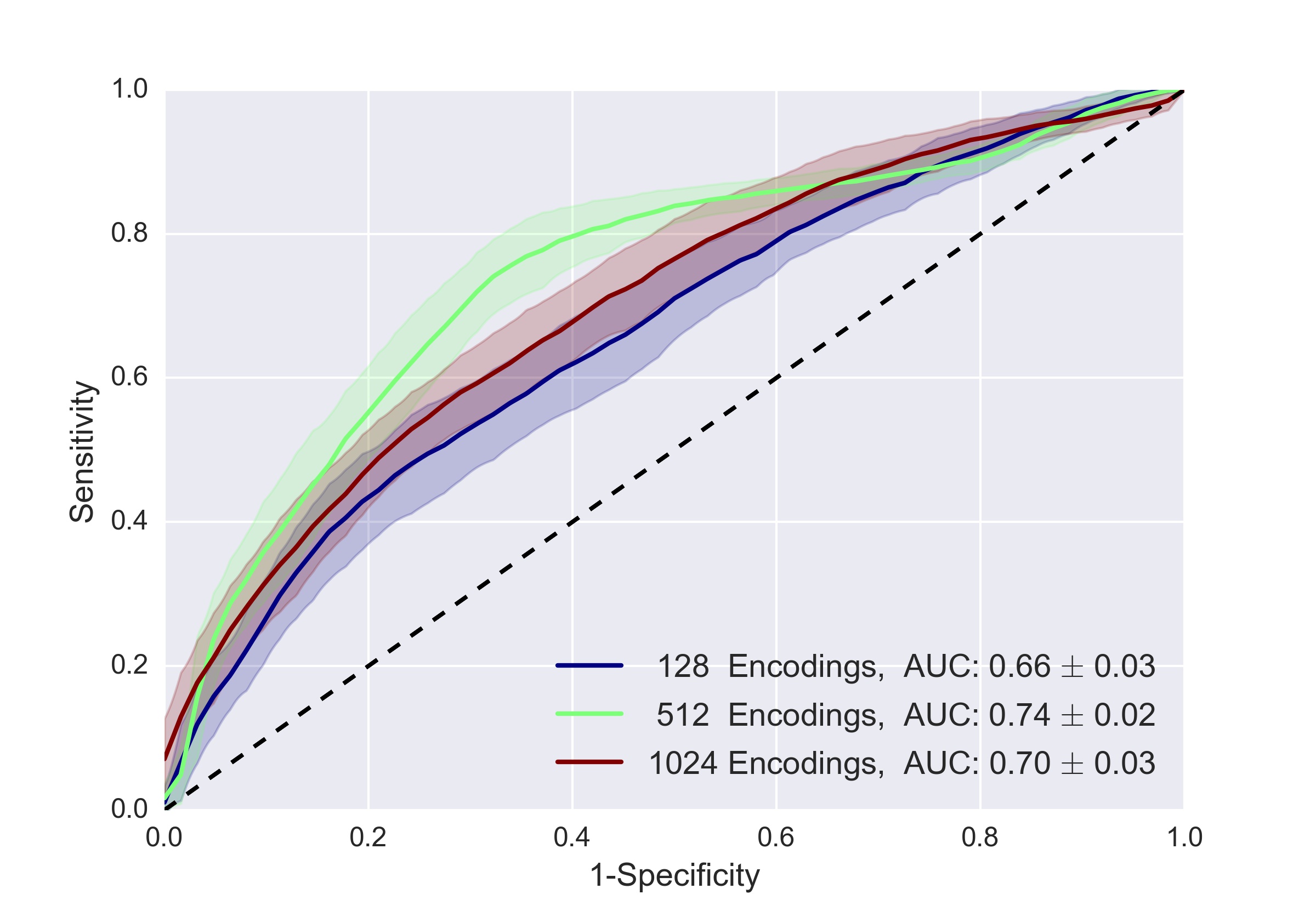}}
		\caption{}
	\end{subfigure}
	\begin{subfigure}{0.45\textwidth}
\centering
		\centerline{\includegraphics[width=1.1\linewidth]{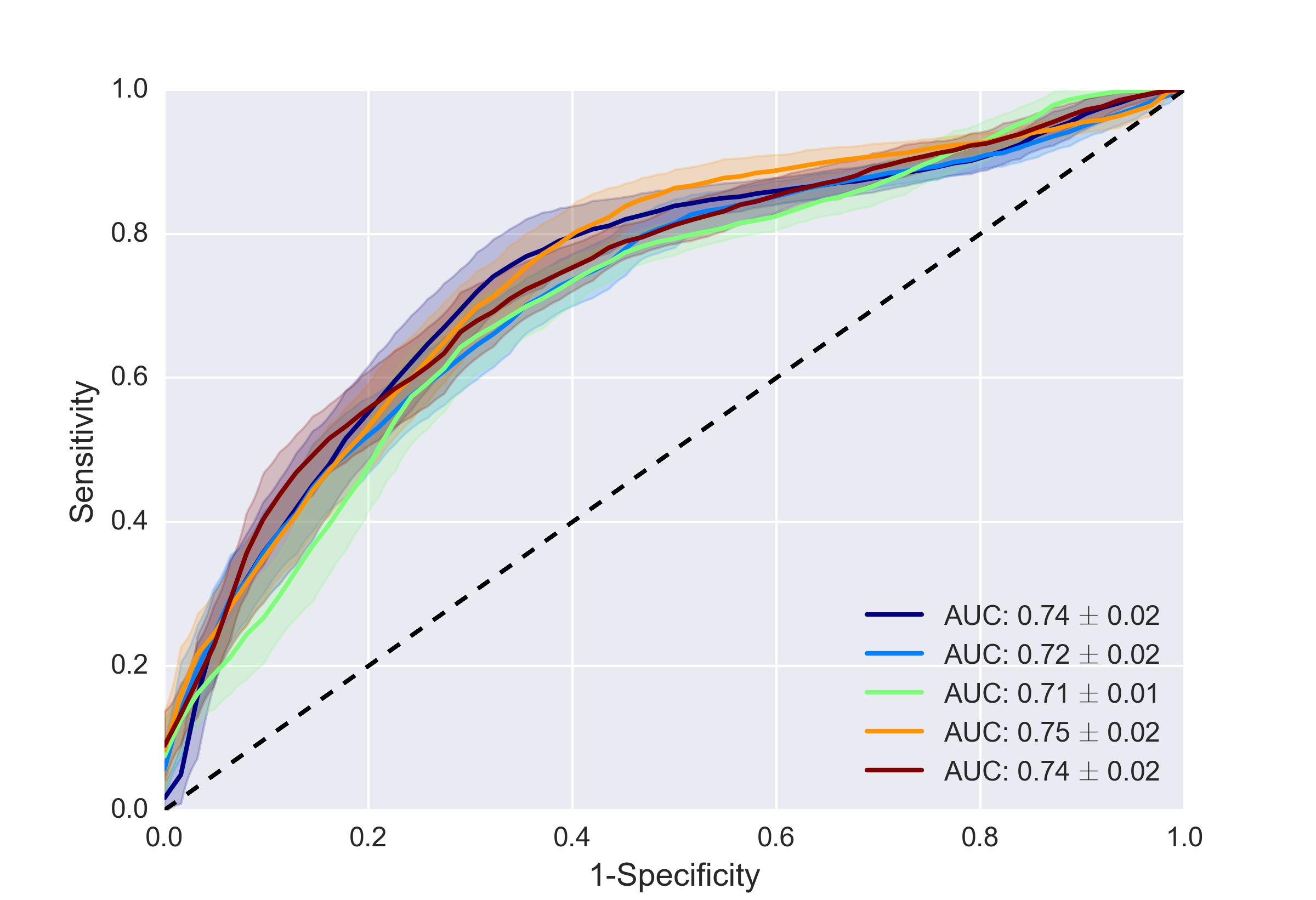}}
		\caption{}
	\end{subfigure}
	\begin{subfigure}{0.45\textwidth}
\centering
	\centerline{\includegraphics[width=1.1\linewidth]{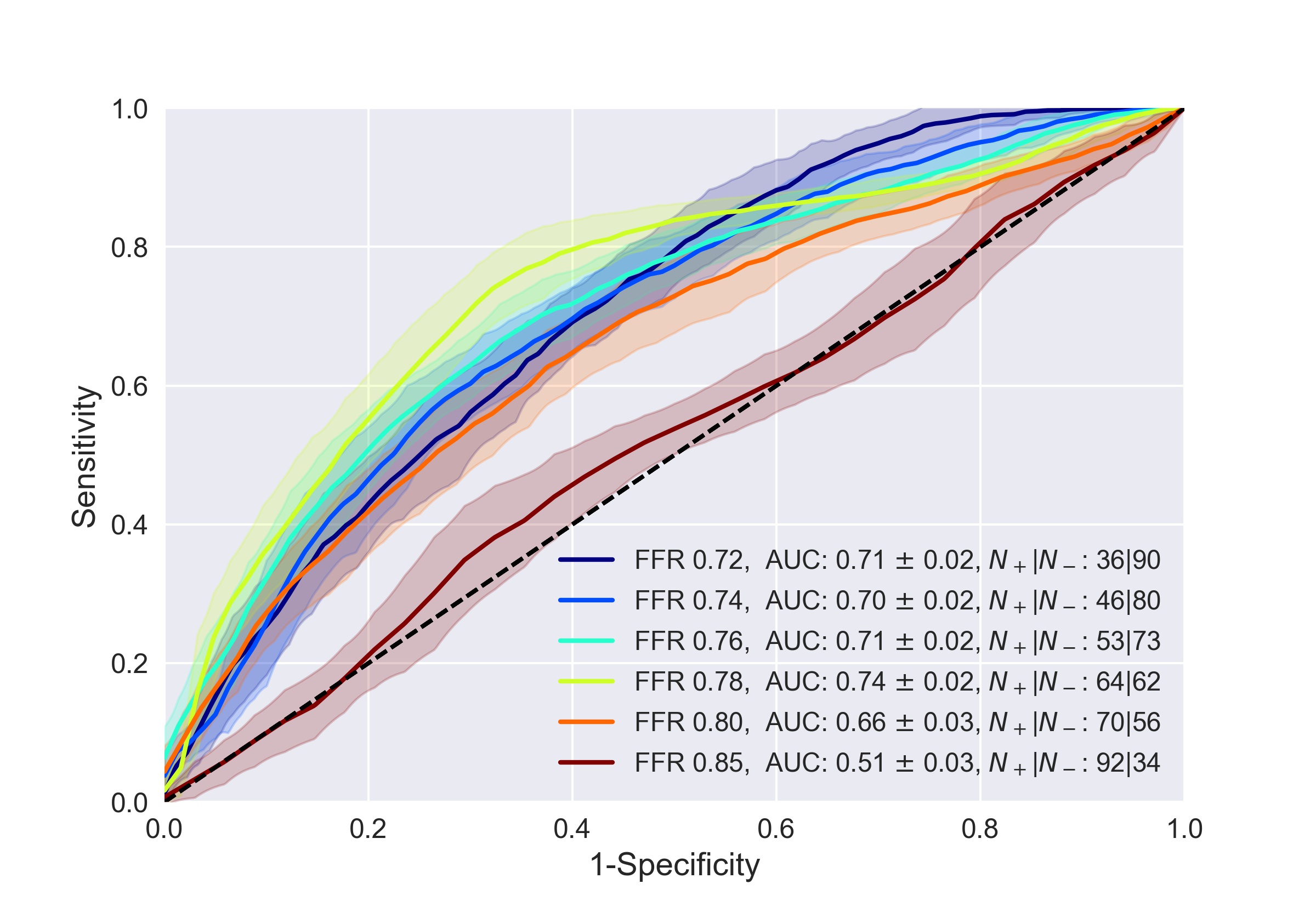}}
	\caption{}
	\end{subfigure}

	\caption{Average ROC curves when (a) varying number of clusters of the LV myocardium with 512 encodings, (b) varying number of encodings of the LV myocardium with 500 clusters, (c) using random seed initialization for clustering of the LV myocardium into 500 clusters with 512 encodings, and (d) using different cut-off values on FFR with 500 clusters and 512 encodings, number of positive and negative samples ($N_{+}|N_{-}$) is shown per cut-off value. In (a), (b) and (c) the cut-off value on the FFR was set to 0.78. The shaded areas show the standard deviations of the sensitivities.}
	
	\label{fig:perf_clusters_encodings_sizes}		 			
\end{figure}

To evaluate the effect of the number of used encodings on patient classification, two additional experiments with different numbers of the encodings, namely 128 and 1024, were performed. \review{For this purpose, two} additional CAEs were trained and used to encode myocardial voxels. The architecture of these CAEs is identical to the one illustrated in Fig. \ref{fig:cae}, only the number of units in the fully connected layer (encodings layer) was changed to 128 or 1024, respectively. In both experiments, the number of clusters was set to 500. The results are illustrated in Fig.~\ref{fig:perf_clusters_encodings_sizes} (b). The \review{obtained} AUC ranged \sout{for different numbers of encodings} from 0.66 to 0.74, while the highest AUC was obtained with 512 encodings.

\subsection{The Effect of Seed Initialization}
The clustering algorithm uses random initialization of the centers of the clusters (seeds) \citep{Scul10}. This may potentially lead to different clusters and thereby different feature values which may lead to different patient classification. To evaluate the effect of the seed initialization, four additional experiments were performed using different random seed initializations. The number of clusters and encodings was not altered, and was set to 500 and 512, respectively. In these experiments, the \review{obtained} AUC ranged from 0.71 to 0.75. The results are shown in Fig.~\ref{fig:perf_clusters_encodings_sizes} (c).

\subsection{The Effect of FFR Cut-off Values}

In the literature \citep{Pijl96,De01,Sama06,De08,Petr13b}, different cut-off values on the FFR have been used to separate patients with functionally significant stenosis from those without it. These values range from 0.72 to 0.8. Therefore, with fixed number of clusters and encodings (500 and 512, respectively), the performance of the classification using five additional FFR cut-off values, namely 0.72, 0.74, 0.76, 0.8 and 0.85, was investigated. The results are shown in Fig.~\ref{fig:perf_clusters_encodings_sizes} (d). %and listed in Table \ref{list:auc_ffr_thr}. 
The \review{obtained} AUC ranged from 0.51 when using the 0.85 cut-off value to 0.74 while using the 0.78 cut-off value on FFR.

\subsection{The Effect of Segmentation Accuracy}\label{seg_sens}

\sout{R2C8} \review{ An accurate LV myocardium segmentation is a prerequisite for accurate patient classification. To evaluate whether inaccuracies generated by our automatic myocardium segmentation have an impact of the patient classification, manual correction of the automatic segmentation results graded as 2 (mostly accurate) or 3 (inaccurate) (Table \ref{list:quality_results}) has been performed. Thereafter, all subsequent analysis using the same set of patients (126) was repeated. The analysis has resulted in an AUC of $0.73\pm 0.02$ vs. AUC $0.74 \pm 0.02$ previously achieved when relying on automatic segmentation. The results were not statistically significant ($p>0.05$).}

\subsection{Comparison with Other Methods}\label{others}

\sout{$R2C1_1$} \review{ To allow comparison of the results obtained by the proposed method with previously published work, Table \ref{list:other_methods_results} lists the results as reported in the original publications. Note that these methods were evaluated with different patients and scans, and by using different evaluation metrics. 
Hence, these results can be used only as indication of the differences in the performance. }

\begin{table}[h!]
	\centering
	\caption{Comparison with previous work. Table lists number of evaluated patients (Patients) and vessels (Vessels), achieved diagnostic accuracy (Accuracy) and the corresponding area under the ROC curve (AUC) per-patient and per-vessel for the detection of functionally significant stenosis as reported in the original studies. Please note that different methods perform either analysis of the blood flow in the coronary arteries (Blood flow) or detect ischemic changes directly in LV myocardium (Myo.). }
\label{list:other_methods_results}

	\resizebox{\textwidth}{!}{%
		\begin{tabular}{cccccccc}
			\hline

			& \multicolumn{1}{l}{} &                                           &                                          & \multicolumn{2}{c}{\textbf{Per-patient}}                                 & \multicolumn{2}{c}{\textbf{Per-vessel}}

			  \\

\cmidrule(lr){5-6}
\cmidrule(lr){7-8}
			
			& \textbf{Study}       & \multicolumn{1}{l}{\textbf{Patients}} & \multicolumn{1}{l}{\textbf{Vessels}} & \multicolumn{1}{l}{\textbf{Accuracy}} & \multicolumn{1}{l}{\textbf{AUC}} & \textbf{Accuracy} & \multicolumn{1}{l}{\textbf{AUC}} \\ \hline
			\multirow{4}{*}{\rotatebox[origin=c]{90}{Blood flow}} & \cite{Min12}         & 252                                       & -                                        & 0.71                                  & 0.81                             & -                 & -                                \\ \cline{2-8} 
			& \cite{Norg14}        & 254                                       & 484                                      & 0.81                                  & 0.90                             & 0.86              & 0.93                             \\ \cline{2-8} 
			& \cite{Renk14}        & 53                                        & 67                                       & 0.86                                  & -                                & 0.85              & -                                \\ \cline{2-8} 
			& \cite{Coen15}        & 106                                       & 189                                      & -                                     & -                                & 0.74              & -                                
			\\ \Xhline{2pt}
			
			\multirow{2}{*}{\rotatebox[origin=c]{90}{Myo.}} & \cite{Han17}         & 252                                       & 407                                      & 0.63                                  & -                                & 0.57              & -                                \\ \cline{2-8} 
			& Ours                 & 126                                       & -                                        & 0.71                                  & 0.74                             & -                 & -                               
		\end{tabular}%
	}
\end{table}

\review{The results indicate that the methods utilizing analysis of the blood flow tend to archive higher per-patient accuracy. The results also show that the proposed approach seems to outperform recently proposed method relying on the LV myocardium analysis proposed by \citet{Han17}, employing the method presented by \citet{Xion15}. Also, the achieved accuracy equal to the method performing analysis of the blood flow by \citet{Min12}.}

\section{Discussion and Conclusion}\label{discussion}

A method for identification of patients with functionally significant stenosis in the coronary arteries has been presented.
Unlike previous methods that determine the functional significance of coronary artery stenosis relying on the analysis of the coronary artery tree in CCTA scans \citep{Tayl13,Stei10,Itu12,Nick15}, this method analyzes the LV myocardium \textit{only}. The algorithm first performs automatic segmentation of the LV myocardium. The segmented myocardium is subsequently encoded using a CAE, and the encoding statistics are used to classify patients into those with and without functionally significant coronary artery stenosis. The reference for the presence of functionally significant coronary artery stenosis was provided by invasively obtained FFR measurements that are currently considered the clinical standard.

Our experiments show that moderate performance \review{with average accuracy, sensitivity and specificity of 0.71, 0.70 and 0.71, respectively} and AUC $0.74 \pm 0.02$ for identifying patients with functionally significant stenosis could be achieved by only using features extracted from the LV myocardium. The results demonstrate that myocardial information, derived directly from a single CCTA at rest using a trained CAE, has reasonable predictive ability compared with invasive FFR measurements obtained during ICA. \sout{$R2C1_2$}\review{As shown in Table \ref{list:other_methods_results}, the proposed method seems to outperform the results reported by \citet{Han17}. Unlike \cite{Han17}, that performed patient classification based on engineered LV myocardium features, we have relied on convolutional auto-encoder to learn the discriminative encodings. %M2R2
	 	\reviewminortwo{ As the characteristics of the LV myocardium in a rest CCTA affected by ischemic changes are not (well-)defined and presently
	\reviewminortwo{there is} no consensus \sout{has been established}regarding their appearance \citep{Spir13,Xion15}, \reviewminortwo{it is} likely \reviewminortwo{that} our unsupervised approach led to higher diagnostic accuracy \reviewminortwo{than the designed features approach presented in} \citet{Han17}}. The learned encodings were able to represent myocardium voxels accurately and densely, as was demonstrated visually by the small mean squared error between the input and the reconstructed examples (Fig.~\ref{fig:reconstructed_patches}). \sout{R3C11}Even though the encodings allowed extraction of features that were used for patient classification, they are hard to interpret precluding visualization of the location and appearance of ischemia.
	In spite of excellent performance of deep learning techniques demonstrated in many medical image analysis tasks, revealing their inner dynamics is beyond the scope of this work. 
}

\sout{ $R1C4_2$ Our findings could be supported by several clinical studies. Busch et al. (2011) showed that myocardial hypo-enhancement, as seen visually by human observers in CCTA images acquired at rest, identified myocardial hypoperfusion with high sensitivity as well as high specificity. Gupta et al. (2013) showed the feasibility of rest CCTA in detecting the presence of infarction and quantifying the perfusion defects compared with standard nuclear myocardial perfusion imaging.}

\sout{R4C8} Segmentation of the LV myocardium is a prerequisite for patient classification. Any accurate automatic (e.g. \cite{Zhen08a,Kiri10,Xion15}) or manual segmentation could be employed. 
In this work, LV myocardium segmentation was performed automatically, and \review{in a limited number of images} the obtained quantitative \sout{and qualitative} evaluation demonstrated high segmentation performance. \sout{R3C9}\review{As patient classification depends on the LV segmentations, a qualitative evaluation of the automatic segmentation was performed on the complete set of CCTA images. Only the images with accurate automatic segmentations of the LV myocardium were analyzed further.}

\sout{R2C7 The achieved quantitative results (Dice $91.4 \pm 2.1 \%$, MAD $0.7 \pm 0.1$ mm) are slightly better than the performance of the second observer's manual annotations (Dice $90.0\pm 2.1\%$, MAD ($0.6\pm 0.1$ mm), which were obtained in our preliminary study on a subset of scans used in this study.}

Nevertheless, automatic segmentation did not achieve satisfactory results in regions where the CCTA images lack contrast between the blood pool and the LV myocardium or between the right ventricle and the LV myocardium. \sout{$R2C8_2$}\review{Therefore, the affect of the accuracy of LV myocardium segmentation on patient classification was examined. The results shown in Section \ref{seg_sens}, indicate that small inaccuracies in the automatic segmentation do not significantly influence patient classification. However, to} further improve the segmentation, future work could take shape priors or constrains into account to compensate for the lack of contrast in these regions. Moreover, fully convolutional networks \citep{Long15} like U-net \citep{Ronn15}, which perform segmentation on an entire image rather than per voxel classification, could be employed to achieve \sout{faster}\review{more accurate} LV myocardium segmentation.\sout{R4C1} \review{ Additionally, in future work, it would be interesting to address the use of 3D analysis instead of 2.5D approach to leverage the volumetric information. However, given the current trade-off between size of the receptive field and hardware limitations, 2.5D approach was advantageous \citep{Hava17}.}

\sout{A trained CAE that provides a number of encodings for the LV myocardial voxels was applied. These encodings were able to represent myocardium voxels accurately, as was demonstrated quantitatively by the small mean squared error between the input and output patches (Fig.~\ref{fig:cae_loss}) and visually by the reconstructed examples (Fig.~\ref{fig:reconstructed_patches})} The number of encodings \review{provided by the CAE} was also investigated (Fig.~\ref{fig:perf_clusters_encodings_sizes}, (b)). Encoding the LV myocardium voxels with fewer encodings proved to have inferior performance \sout{compared to 512 encodings}. This is likely because of aggressive compression by the encoder, which could lead to underrepresentation of the input texture and morphology contained in the encodings, and therefore loosing valuable information. Increasing the number of encodings proved to be non-beneficial either, as in this case, the CAE possibly overfits the input patches due to its massive number of hyperparameters. 

The myocardium was spatially clustered to preserve the local nature of ischemic changes in the myocardial texture. Size \review{of the clusters}, controlled by their number\sout{of clusters}, is likely important for identification of patients with ischemia. Therefore, the number of defined clusters was investigated (Fig.~\ref{fig:perf_clusters_encodings_sizes}, (a), Section \ref{clustersresults}). As expected, analyzing the myocardium as one cluster resulted in the worst classification performance. This is likely due to local impact of ischemia which is typically confined to the myocardial territory perfused by the stenotic coronary artery \citep{Mehr11,Ross14}. Thus, analyzing the myocardium globally (one cluster) masked local changes. \sout{$R1C3_2$}\review{ Moreover, clustering the myocardium into small number of clusters (e.g. 17 like in the AHA 17-segment model or 10 as in Fig.~\ref{fig:perf_clusters_encodings_sizes}, (a)) has not shown good performance, likely for the same reason. However,}
using too many clusters was not beneficial either, since the size of each cluster was likely too small to capture the inhomogeneity within that region.

Additional evaluation was performed using random but different seed initializations used for clustering the LV myocardium. The results (Fig.~\ref{fig:perf_clusters_encodings_sizes} (c)) show that the method is sensitive to initialization, where the average AUC for patient classification ranged from 0.71 to 0.75. These findings suggest that further research towards finding a more robust clustering approach needs to be conducted. \sout{$R1C3_3$}\review{ For instance, dividing the LV myocardium into territories that enable relating them with the coronary arteries segments perfusing them could be beneficial, as has been proven in the visual assessment of myocardial perfusion defects \citep{Cerc12}.} \sout{clustering the LV myocardium while aligning the myocardial clusters to the coronary artery segments and to myocardial perfusion territories has been proven beneficial for visual assessment of myocardial perfusion defects in CCTA }However, this would require automatic localization of several anatomical landmarks, like the mitral valve and the apex of the heart, which is beyond the scope of the current work. Moreover, clustering the LV myocardium into supervoxels, using e.g SLIC \citep{Acha12}, and taking into account the values of the myocardial voxels rather than their locations, might be beneficial. 

In this work, simple statistics, namely standard deviation and maximum, were used to reduce the high dimensionality of the clustered voxels' encodings. Future work could investigate alternative ways of compressing the local encodings, e.g. restricted Boltzmann machine or deep belief networks \citep{Hint06,Lee09_2,Beng13}. These generative approaches, which belong to the undirected graphical models, could be employed to represent a group of voxel encodings by more compressed but yet descriptive representations. Moreover, the CAE network was restricted to the analysis of 2D patches to avoid overfitting, to keep the number of trainable hyperparameters low, and to limit computational costs. Nonetheless, encoding 3D sub-volumes of the LV myocardium could be beneficial because the myocardium is a 3D object and extracting features from its volumetric sub-regions could capture useful information that may have not been captured in the 2D encoding. 

Several studies investigated the reproducibility and the variability of repeated FFR measurements \citep{Ntal10,Berr13,Petr13b,John14}. These studies reported different values (range $3\%$ to $5\%$) for the standard deviation of the differences in repeated FFR measurements. However, these studies all agree that when a measured FFR falls within the "\textit{gray-zone}", a value between 0.75 and 0.80, a single fixed cut-off point on the FFR measurement cannot by itself determine the functional significance of a stenosis, and therefore a broad clinical judgment is needed. Within this gray-zone, the uncertainty of an FFR measurement is high. In this study, we also evaluated different cut-off points on FFR (Fig.~\ref{fig:perf_clusters_encodings_sizes} (d)). When a high value of 0.85 and a low value of 0.72 were used, the data set was unbalanced with respect to class labels. Namely, when a value of 0.85 was used for the FFR cut-off point, the dataset contained 92 positive and 34 negative samples, and when 0.72 was used, the dataset contained 36 positive and 90 negative samples (Fig.~\ref{fig:perf_clusters_encodings_sizes} (d)). While these two cases are almost symmetrical with respect to the positive and negative class sample ration, the method only performs poorly (AUC $0.51 \pm 0.03$) when using 0.85 as a cut-off value. This result was expected as this cut-off point is not hemodynamically or functionally relevant and is not used in clinical practice to identify functionally significant stenoses. These results show that patient classification is not biased towards the majority class in the dataset, but is sensitive to the functional significance of the stenoses.

Assessment of the coronary artery tree in CCTA scans has a high sensitivity in detecting the functional significance of a detected stenosis, but with limited specificity \citep{Meij08,Bamb11,Ko12}. Therefore, to improve the specificity of CCTA and prevent patients from undergoing ICA unnecessarily, future work will investigate the incremental diagnostic value of integrating the presented method with the assessment of the coronary arteries and their characteristics in CCTA. 

\review{ The proposed approach does not identify specific stenosis that is functionally significant but only determines whether a patient has a stenosis that is functionally significant. Future work may address this limitation}. %Rmonor
\reviewminor{Given the moderate accuracy of our and the previous method \citep{Han17}, future studies are needed to increase the accuracy of these approaches. For this purpose, a combination of computational fluid dynamics in the coronary arteries and image analysis of the LV myocardium could be considered.
Finally, future work should evaluate this method using a larger set of scans from different vendors and hospitals.}

To conclude, this study presented a novel algorithm for automatic classification of patients according to the presence of functionally significant stenosis in one or more coronary arteries, using only information extracted from the LV myocardium in a single rest CCTA scan. \sout{$R2C4_2$}\review{This could potentially reduce the number of patients that unnecessarily undergo invasive FFR measurements.}

\section{Acknowledgments}

We would like to thank Samuel St-Jean for helping us out with \LaTeX.

This study was financially supported by the project FSCAD, funded by the Netherlands Organization for Health Research and Development (ZonMw) in the framework of the research programme IMDI (Innovative Medical Devices Initiative); project 104003009.
We gratefully acknowledge the support of NVIDIA Corporation
with the donation of the Tesla K40 GPU used for this research.

%\section*{References}

%\bibliography{D:/Literature/CAD}
\bibliography{CAD}
\end{document}